%% file: main.tex
\documentclass[10pt,twocolumn,letterpaper]{article}
\usepackage[T1]{fontenc}   
\usepackage[pagenumbers]{cvpr} 
\input{preamble}
\input{macros}

\definecolor{cvprblue}{rgb}{0.21,0.49,0.74}
\usepackage[pagebackref,breaklinks,colorlinks,allcolors=cvprblue]{hyperref}

\def\confName{CVPR}
\def\confYear{2027}

\title{\method: Fast, Verified Memory for Agentic Vision-Language-Action Policies}

\author{
Jike Zhong\textsuperscript{1}\quad
Yuxiang Lai\textsuperscript{2}\quad
Ke Wang\textsuperscript{3}\quad
Ming Li\textsuperscript{4}\quad
Qianzhong Chen\textsuperscript{3}\quad
Xi Xiao\textsuperscript{5}\\
Ethan Y. Don\textsuperscript{1}\quad
Arpita Chowdhury\textsuperscript{6}\quad
Shao-Yuan Lo\textsuperscript{7}\quad
Konstantinos Psounis\textsuperscript{1}\quad
Zheda Mai\textsuperscript{8}\\[4pt]
\textsuperscript{1}University of Southern California\quad
\textsuperscript{2}Emory University\quad
\textsuperscript{3}Stanford University\\
\textsuperscript{4}University of Florida\quad
\textsuperscript{5}University of Alabama at Birmingham\quad
\textsuperscript{6}Boston University\\
\textsuperscript{7}National Taiwan University\quad
\textsuperscript{8}The Ohio State University
}

\begin{document}
\maketitle
\input{fig/teaser}
\input{sec/0_abstract}
\input{sec/1_intro}
\input{sec/2_related}
\input{sec/3_method}
\input{sec/4_experiments}
\input{sec/5_conclusion}
\input{sec/6_llm_statement}
{
    \small
    \bibliographystyle{ieeenat_fullname}
    \bibliography{main}
}

\makeatletter\let\Hy@backout\@gobble\makeatother 
\input{sec/X_suppl}

\end{document}

%% file: preamble.tex
\usepackage{cuted} 

\usepackage{currfile} 

\usepackage{caption} 
\usepackage{microtype}



%% file: macros.tex
\usepackage{multirow}
\usepackage{colortbl}
\usepackage{tabularx}
\usepackage{pifont}
\usepackage{xspace}
\usepackage{array}
\usepackage{makecell}
\usepackage{placeins}
\graphicspath{{figures/}}

\newcommand{\method}{ReflexMem\xspace}
\newcommand{\pih}{$\pi_{0.5}$\xspace}
\newcommand{\robomme}{RoboMME\xspace}
\newcommand{\rmbench}{RMBench\xspace}

\definecolor{ours}{HTML}{0F4D92}
\definecolor{oursbg}{HTML}{E4ECF6}
\definecolor{writec}{HTML}{2E8B57}
\definecolor{memc}{HTML}{B07A00}
\definecolor{readc}{HTML}{9A4D8E}
\definecolor{refgray}{HTML}{8A8A8A}

\newcommand{\best}[1]{\textbf{#1}}
\newcommand{\second}[1]{\underline{#1}}
\newcommand{\refrow}[1]{\textcolor{refgray}{#1}}
\newcommand{\oursrow}{\rowcolor{oursbg}}

\newcommand{\rep}{$^{\dagger}$}

\renewcommand{\paragraph}[1]{\vspace{.35em}\noindent\textbf{#1.}}

%% file: fig/teaser.tex
\begin{strip}
\centering
\vspace{-2em}\vspace{-8pt}
\includegraphics[width=\linewidth]{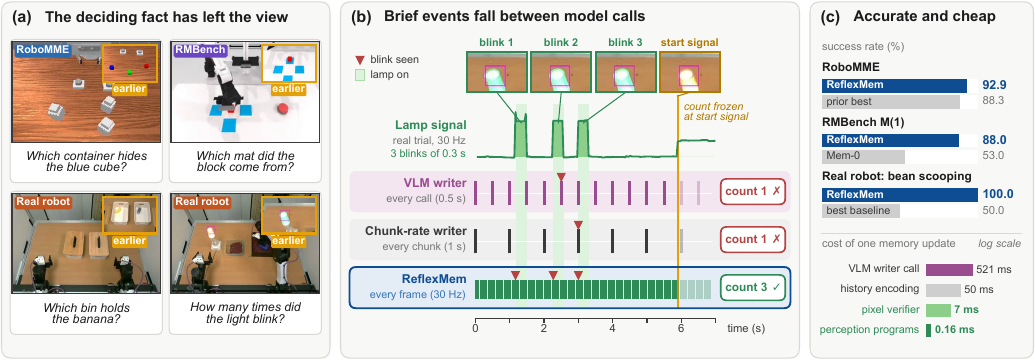}
\captionsetup{hypcap=false}
\captionof{figure}{\textbf{Memory should be written at the rate of the world and read only when it changes.} (a) In memory-dependent manipulation, the fact that decides the next action is no longer visible when the robot acts (insets: the earlier moment at which it was). (b) Lamp signal of a real trial with three blinks and the instants at which three write schedules observe it ($\blacktriangledown$: a blink seen by the writer; pink box: lamp region): writing once per VLM call or once per action chunk misses blinks, whereas \method writes from every frame. (c) Success rates of \method against the best published result on \robomme, the released Mem-0 on the single-fact (M(1)) tasks of \rmbench, and the strongest of four baselines on real-robot bean scooping; measured cost of one memory update (perception programs: per camera frame of the real-robot task; history encoding as in~\cite{robomme}).}
\captionsetup{hypcap=true}
\label{fig:teaser}
\vspace{-0.4em}
\end{strip}

%% file: sec/0_abstract.tex
\begin{abstract}
Vision-language-action (VLA) policies act on the current observation, yet many manipulation tasks hinge on facts that are no longer visible: how many times a lamp blinked, which bin holds an object, whether the last placement succeeded. Existing memory-augmented VLAs either compress visual history into tokens that the policy must learn to decode, or let a vision-language model (VLM) write the memory once per decision, which can miss brief events between calls and may record what was commanded rather than what happened. Our key insight is that writing a memory and reading it have different requirements: writing must be fast and verified, while reading needs reasoning only when the memory changes. To this end, we propose \textbf{\method}, a memory layer for agentic VLA systems, in which a language-level planner steers a VLA policy. Lightweight perception programs process the observation stream on a CPU and detect candidate events; a verifier confirms events caused by the robot before they are written to an explicit, human-readable task state; and a subgoal writer is invoked when this state changes rather than at every decision, passing a short instruction for the next step to a VLA policy whose architecture is unchanged. With \pih as the policy, \method completes 60 of 60 trials on two real-robot tasks, twice the strongest baseline on bean scooping, and achieves 92.9\% on \robomme with agent-written motion routines, within about three points of ground-truth memory, and 88.0\% on the single-fact tasks of \rmbench. The write path costs as little as 0.16\,ms per camera frame, orders of magnitude below a VLM call.
\end{abstract}

%% file: sec/1_intro.tex
\section{Introduction}
\label{sec:intro}

Vision-language-action (VLA) models~\cite{rt2,openvla,octo,pi0,pi05} are rapidly becoming the default interface for general-purpose robot manipulation. Most are reactive: each chunk of actions is predicted from the current camera images and the instruction alone. A robot asked to scoop beans ``as many times as the light blinked'' must act on a count that is no longer visible; a robot asked for the banana must recall which of two closed bins holds it. Benchmarks for such \emph{memory-dependent} manipulation quantify the gap: the reactive \pih~\cite{pi05} succeeds in only 17.9\% of the episodes of \robomme~\cite{robomme} and in 14.4\% on the tasks of \rmbench~\cite{rmbench} that we study.

Two families of remedies have emerged (\cref{sec:related}). \emph{Perceptual} memories feed sampled frames, compressed tokens or video context into the policy~\cite{memoryvla,eventvla,evomem,simplememvla}; the facts that decide the next action (the \emph{task state}) stay implicit and are re-encoded at every decision. \emph{Symbolic} memories keep the task state outside the policy, written by a vision-language model (VLM)~\cite{robomme,memer} or with perception tools around a VLM or LLM planner~\cite{ledger,simplearm,hymes}; they are readable, but they keep a large model in the write path or call one on a fixed schedule, with two consequences. First, a model called once per decision sees only a few sampled frames between calls, so brief events, such as a 0.3\,s lamp blink, are lost (\cref{fig:teaser}b). Second, what is written is often the commanded subgoal~\cite{robomme,evomem}, and commanded actions fail: a missed press that is remembered as a press corrupts every later decision.

Our key insight is that a memory has two operations with different requirements. \emph{Writing} must keep pace with the world and must be backed by evidence that an event actually occurred. \emph{Reading}, that is, reasoning over the remembered state to decide what to do next, can use the flexibility of a language model, but needs to run only when the state changes. A memory should therefore run on two clocks.

To this end, we present \textbf{\method}, a memory layer built on this principle for agentic VLA systems, in which a language-level planner steers a VLA policy (\cref{fig:pipeline}). Memory is written like a reflex and read like a plan. (1) \textbf{Perception programs}, short image-processing and kinematic routines, process the observation stream on a CPU and raise candidate events. (2) \textbf{Verified commits}: an event that the robot's own action should cause is written to an explicit, human-readable task state only after a learned pixel verifier, or a test on measured joint states, confirms it. (3) A \textbf{state-triggered subgoal writer}, a template or a VLM, is called when the task state changes rather than at every decision, and hands the next \emph{language subgoal}, a short instruction for the next step, to the policy through its prompt.

With \pih as the policy, \method completes 60 of 60 trials of two memory-dependent tasks on a real bimanual robot, where the policy executes every action and no VLM is called; on bean scooping the strongest of four baselines reaches 50\%. On \robomme it achieves a 92.9\% success rate, 4.6 points above the best published result, and on the five single-fact (M(1)) tasks of \rmbench 88.0\%, where the benchmark's own memory policy, Mem-0, reaches 53.0\% on the same seeds. On the eleven \robomme tasks that the released policy executes (a motion routine acts on the other five), \method reaches 93.3\%, against 90.2\% for the best published method. Removing the task state reduces the success rate on \rmbench from 88.0\% to 26.0\%; committing on a timer instead of the verifier reduces it on \robomme from 80.9\% to 47.0\%; and at one write per action chunk, even an ideal per-frame detector recovers the blink count in only 8.6\% of cases. Our contributions are as follows:
\begin{itemize}
    \item We identify the coupling of memory writing to the rate of a large model as a limitation of symbolic memories for VLAs, and propose to separate fast, verified \emph{writing} from state-triggered \emph{reading}.
    \item We introduce \textbf{\method}, which realizes this separation with perception programs, verified commits to an explicit task state, and a state-triggered subgoal writer around a VLA policy whose architecture is unchanged.
    \item We evaluate \method on two benchmarks and two real-robot tasks, with memory-removed controls, ablations of verification and write rate, configuration by a fresh coding agent (94.4\% on 750 \robomme test episodes), and a same-hardware cost comparison.
\end{itemize}

%% file: sec/2_related.tex
\section{Related work}
\label{sec:related}

\begin{figure*}[t]
    \centering
    \includegraphics[width=\linewidth]{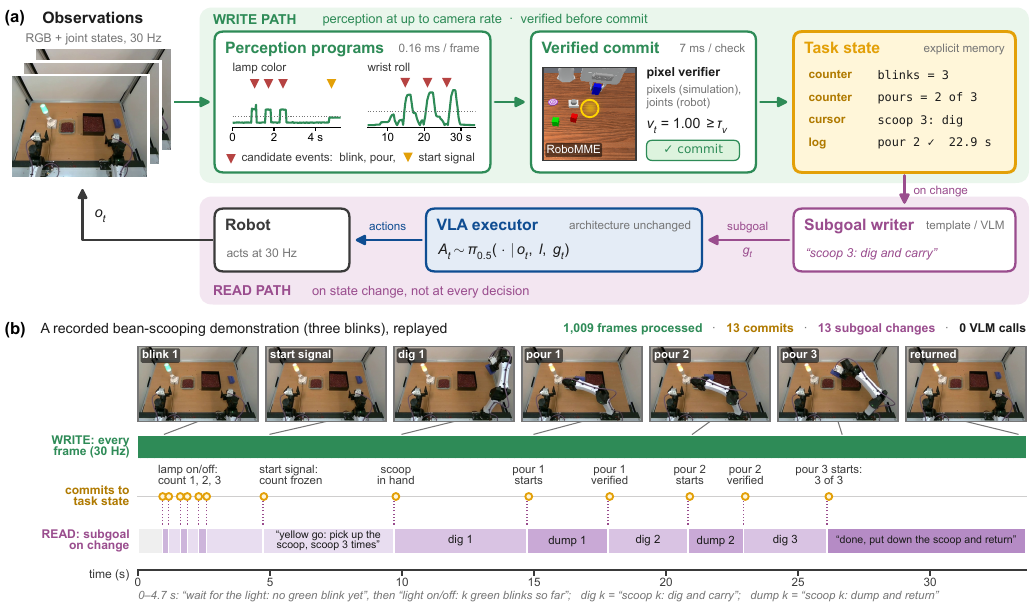}
    \caption{\textbf{Overview of \method.} (a) The \textcolor{writec}{write path} runs lightweight perception programs on the observation stream; events caused by the robot are committed to the explicit \textcolor{memc}{task state} only after verification (\cref{sec:write}). The \textcolor{readc}{read path} calls the subgoal writer when the task state changes and hands a language subgoal to the \pih policy through its prompt (\cref{sec:read}). (b) A recorded bean-scooping demonstration with three blinks, replayed through both paths: the perception programs process all 1{,}009 frames, 13 events are committed up to the final subgoal, and the subgoal sent to the policy changes only at these commits. Signals, events and subgoals are recorded data; the verifier example is from \robomme.}
    \label{fig:pipeline}
\vspace{-10pt}
\end{figure*}

\paragraph{Vision-language-action models}
VLA models map images and language to robot actions~\cite{rt2,openvla,pi0,pi05}. Recent work adds reasoning to this reactive core: visual subgoals~\cite{cotvla}, latent reasoning tokens~\cite{fastthinkact}, predicted affordances and progress~\cite{palm}, and subtask planners~\cite{foreact,robobrain}. \method instead supplies the state that the current observation lacks.

\paragraph{Memory for robot manipulation}
\robomme~\cite{robomme} and \rmbench~\cite{rmbench} isolate tasks whose correct action depends on the past (see also~\cite{mikasa,robomemarena}). One family of methods stores the past \emph{inside} the policy, as sampled frames, token banks, keyframes, video context or past reasoning~\cite{robomme,memoryvla,eventvla,evomem,simplememvla,hifvla,optimusvla,trmvla}; these memories remain implicit. A second family keeps the state \emph{outside} the policy. A VLM emits the next subgoal from its own past subgoals~\cite{robomme} or from keyframes it selects at about 1\,Hz~\cite{memer}, or routes a latent summary of its context to the action model~\cite{ponderpounce}; Statler lets one language model write a world state from the commanded action and another read it~\cite{statler}. Recent systems maintain the state with perception tools: Ledger runs a tracker and calls an LLM planner at step boundaries~\cite{ledger}, SimpleARM gates a typed state on tool evidence while its subgoal VLM is queried every 16 steps~\cite{simplearm}, and HyMeS advances a state held in generated code by votes over periodic VLM and proprioceptive queries~\cite{hymes}; Mem-0 and Notes-to-Self update on a trigger that the model itself declares~\cite{rmbench,notestoself}. Each keeps a large model in its write path or calls one on a fixed schedule. In \method the write path is a set of CPU programs that run at up to camera rate, robot-caused commits pass a verifier of the commanded outcome, and the subgoal writer is called on any change of the state, by the robot or the environment.

\paragraph{Perception programs and verification}
Language models plan and program robots~\cite{sayCan,codeaspolicies,innermonologue}, and vision routines composed into programs answer visual queries~\cite{visprog,vipergpt}; \method runs such programs as always-on memory writers whose outputs are events rather than answers. Whether a commanded step has happened is the question of success detectors and execution monitors: VLM success detection~\cite{successvqa}, constraints monitored by a VLM~\cite{doremi} or by generated visual programs~\cite{codeasmonitor}, summaries of experience that explain failures~\cite{reflect} and, within memory systems, subtask-end classifiers, proprioceptive checks, VLM queries and progress estimates~\cite{rmbench,ledger,hymes,foreact,r2vlm,palm}. These detect the failures of a plan; \method makes the check the gate of memory writes, decided by a small target-conditioned pixel verifier or by measured joint states, so that the task state records outcomes. Layered robot architectures have long paired a reactive layer with a deliberative one~\cite{threet}; \method brings this split to the memory of a learned VLA policy, written from pixels and read by a language model.

%% file: sec/3_method.tex
\section{Method}
\label{sec:method}

We formalize memory-dependent manipulation (\cref{sec:formulation}) and then describe the write path (\cref{sec:write}) and the read path (\cref{sec:read}) of \method; \cref{fig:pipeline} gives an overview.

\subsection{Problem formulation}
\label{sec:formulation}

\paragraph{Reactive VLA policies}
Given a language instruction $l$ and the current observation $o_t$, which consists of camera images and the proprioceptive state, a VLA policy $\pi_\theta$ predicts a chunk of $H$ future actions,
\begin{equation}
    A_t = [a_t, \dots, a_{t+H-1}] \sim \pi_\theta(\,\cdot \mid o_t, l\,),
    \label{eq:vla}
\end{equation}
executes the first $h \le H$ of them, and is queried again; we call each re-query a \emph{decision} and its time a \emph{chunk boundary}. The policy is \emph{reactive}: nothing observed before time $t$ reaches it. Throughout, $\pi_\theta$ (the \emph{executor}) is \pih~\cite{pi05}.

\paragraph{Memory-dependent tasks}
In a memory-dependent task the correct action is not a function of $(o_t, l)$ alone: two episodes can reach the same observation and require different actions, because a light blinked a different number of times or a different object was shown before the scene was covered. We call the information that separates such episodes the \emph{task state} $s_t$. As in the symbolic-memory baselines of \robomme~\cite{robomme}, the task state reaches the policy as a \emph{language subgoal} $g_t$, a short instruction for the next step that is appended to the prompt and is \emph{grounded} when it carries the image coordinates of its target; we call $\mathcal{W}$ the \emph{write} and $\mathcal{R}$ the \emph{read} operation of the memory:
\begin{align}
    s_t &= \mathcal{W}(s_{t-1},\, o_{\le t},\, g_{t-1}), \label{eq:write}\\
    g_t &= \mathcal{R}(l,\, s_t,\, o_t), \label{eq:read}\\
    A_t &\sim \pi_\theta(\,\cdot \mid o_t, l, g_t\,). \label{eq:act}
\end{align}

\paragraph{Two operations, two rates}
Symbolic memories written by a VLM implement \cref{eq:write,eq:read} with one model call per decision~\cite{robomme,memer}, so both operations run at the rate of that model. Their requirements differ. $\mathcal{W}$ must observe the scene more often than its briefest event lasts, and it must be \emph{faithful}: $s_t$ should record what happened, which is not always what the policy was asked to do. $\mathcal{R}$ can use a language model and needs to be re-evaluated only when $s_t$ changes. \method therefore implements $\mathcal{W}$ with perception programs on the observation stream and a verifier at every chunk boundary, and invokes $\mathcal{R}$ when $s_t$ changes.

\subsection{Write path: fast perception, verified commits}
\label{sec:write}

\paragraph{Perception programs}
The write path starts from a small library of \emph{perception programs}. A program $\phi_k$ is a short image-processing or kinematic routine that maps the current observation either to a scalar signal $z^k_t = \phi_k(o_t)$, such as the color excess in a region around a lamp or the roll angle of a wrist, or to a \emph{fact} about the scene, such as the centroid and color of each object on the table or whether a tracked object lies inside a container. Programs run on a CPU: on every camera frame on the robot, every 2--8 environment steps in \robomme and every 10 actions in \rmbench. A signal raises a \emph{candidate event} through a detector with hysteresis: with detector state $u^k_t \in \{0,1\}$, $u^k_0 = 0$,
\begin{equation}
    u^k_t =
    \begin{cases}
        1 & \text{if } z^k_t > \tau^k_{\mathrm{on}},\\
        0 & \text{if } z^k_t < \tau^k_{\mathrm{off}},\\
        u^k_{t-1} & \text{otherwise},
    \end{cases}
    \label{eq:event}
\end{equation}
where $\tau^k_{\mathrm{off}} < \tau^k_{\mathrm{on}}$. An \emph{onset} event fires when $u^k$ switches from 0 to 1 (``the green lamp turned on'') and a \emph{return} event when it switches back. A task's configuration is written by an LLM coding agent, which selects programs and sets their regions and thresholds on training episodes; none of it is written or tuned by hand.

\paragraph{Task state}
The task state $s_t$ is a typed, human-readable record with four kinds of entries: \emph{counters} (``2 of 3 pours completed''), \emph{bindings} from objects to locations (``banana $\rightarrow$ left bin''), a \emph{cursor} into the ordered steps of the instruction, and a \emph{log} of committed events with their timestamps. Committing an event $e$ applies its update to the record, for example incrementing a counter or advancing the cursor; we write $s_t = s_{t-1} \oplus e$.

\paragraph{Verified commits}
Events caused by the environment, such as a lamp or an operator rearranging the scene, are committed as soon as their detector fires. Other events are outcomes of the robot's own action; the pending event $e$ is then ``the current subgoal $g_{t-1}$ is complete''. Commanded actions fail (a grasp slips, a press misses), and a memory that records such an event when commanded inherits every execution failure. \method commits a robot-caused event only after verification,
\begin{equation}
    s_t = \begin{cases}
        s_{t-1} \oplus e & \text{if } v_t \ge \tau_v,\\
        s_{t-1} & \text{otherwise},
    \end{cases}
    \label{eq:commit}
\end{equation}
where $v_t \in [0,1]$ scores the evidence that $e$ has occurred. A rejected event stays pending while the policy keeps acting and is scored again at the next chunk boundary.

\begin{figure}[t]
    \centering
    \includegraphics[width=\linewidth]{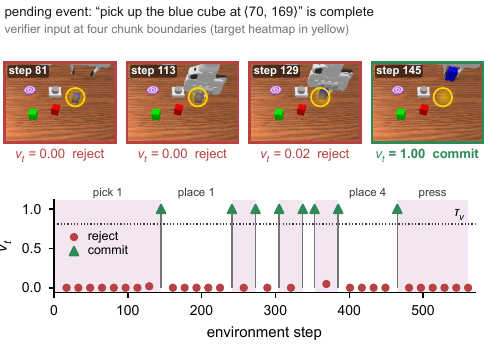}
    \caption{\textbf{Verified commits on a \robomme episode.} Top: the pending event is scored at successive chunk boundaries and committed once the cube has been picked up. Bottom: score $v_t$ at every chunk boundary; bands mark successive subgoals.}
    \label{fig:verifier}
\end{figure}

When completion must be read from the image, as on \robomme, $v_t$ is produced by a learned \emph{pixel verifier} (\cref{fig:verifier}). It sees the front and wrist images at the start of the subgoal and at the three latest chunk boundaries, the robot state, and the subgoal's target, drawn as Gaussian heatmaps in two extra image channels. A ResNet-18~\cite{resnet} encodes each image into one token, and a two-layer transformer pools the eight image tokens with tokens for the robot state, the task, the verb of the subgoal and the elapsed time to predict $v_t$. The verifier is trained with a binary cross-entropy loss on rollouts of the executor in the training scenes, for which the simulator provides the completion step of each subgoal. When completion is defined by the robot's own motion, as for a release, a placement or a pour on \rmbench and on the real robot, the evidence is the \emph{measured} joint trajectory: a pour is committed after the measured wrist roll has left and re-entered its rest band, never because a pour was commanded. Actions that neither source can confirm are logged as attempted.

\subsection{Read path: reasoning on state change}
\label{sec:read}

\paragraph{State-triggered subgoal writer}
The read operation of \cref{eq:read} is executed by a \emph{subgoal writer} that is called only when a commit changes the task state or when the current subgoal exceeds a fixed budget of actions. Where the next subgoal follows from the task state alone, the writer is a template filled from the record. Otherwise it is a VLM, Qwen3-VL-4B~\cite{qwen3vl} with a low-rank adapter~\cite{lora}, prompted with the instruction, the facts from the perception programs, the log of committed events and the current frame. The output is a language subgoal such as ``press the second button at $\langle 60, 105\rangle$'' or ``scoop 2: dig and carry''. Because the writer is state-triggered, the number of VLM calls scales with the number of task steps rather than with the number of action chunks.

\paragraph{Executor interface}
The subgoal is appended to the instruction in the prompt of the \pih policy (\cref{eq:act}); the architecture of the policy is left unchanged. Some remembered facts are spatial references that language expresses poorly, such as ``the object that was on the shelf'' or the original pose of a part. For these the writer additionally renders the fact into the policy's camera image as a box or an outline (a \emph{rendered cue}), so that memory reaches the executor through its two native inputs, text and pixels.

\paragraph{Subgoal-conditioned fine-tuning}
When we fine-tune the executor ourselves, every training frame carries its subgoal in the prompt, and frames around subgoal switches are sampled more often, since there the same observation must lead to different actions.

%% file: sec/4_experiments.tex
\section{Experiments}
\label{sec:exp}

We ask four questions. \textbf{Q1:} How does \method compare with perceptual and symbolic memories on benchmarks (\cref{sec:results})? \textbf{Q2:} Does it transfer to a real robot (\cref{sec:real})? \textbf{Q3:} Which ingredients matter (\cref{sec:ablation})? \textbf{Q4:} What does the memory cost at run time (\cref{sec:efficiency})?

\subsection{Experimental setup}
\label{sec:setup}

\paragraph{Benchmarks}
\robomme~\cite{robomme} contains 16 tabletop tasks in four suites (Counting, Permanence, Reference, Imitation) on a Franka arm with a front and a wrist camera. We use its test protocol: 50 episodes per task with the benchmark's seeds (800 episodes) and 20-action chunks of which 16 are executed. \rmbench~\cite{rmbench} contains nine bimanual tasks; we evaluate the five of its M(1) tier, in which one fact must be retained, on 10 fixed test seeds each.

\paragraph{Real robot}
The robot is a dual-arm YAM (14-D state and action) with a top camera and two wrist cameras. Demonstrations are teleoperated with GELLO~\cite{gello}; every method commands 30-action chunks at 30\,Hz. In \emph{hidden-object recall}, objects are shown in open lidded bins, the lids are closed, and the robot is asked for one object and must open its bin; we run it with a banana and a gray box in two bins (\emph{two-bin recall}). In \emph{bean scooping}, the instruction is ``scoop the beans into the tray as many times as the green light blinked'': a green lamp flashes one to three times, a yellow lamp then gives the start signal, and the robot must scoop and pour exactly that many times and put the scoop down.

\begin{table*}[t]
\centering
\caption{\textbf{Results on \robomme} (success rate, \%). Baseline rows are means over the runs reported in each paper (50 episodes per task; 100 for EvoMem-VLA); \rep{} marks our run of the released checkpoint. \method uses the benchmark's released \pih checkpoint as executor, with an agent-written motion routine on nine tasks (\cref{sec:setup}). Gray rows are references that are not ranked. \textbf{Bold}: best; \underline{underline}: second best.}
\label{tab:robomme}
\small
\setlength{\tabcolsep}{7pt}
\begin{tabular}{@{}llccccc@{}}
\toprule
Method & Memory & Counting & Permanence & Reference & Imitation & Avg. \\
\midrule
\refrow{Human~\cite{robomme}} & \refrow{--} & \refrow{88.5} & \refrow{91.0} & \refrow{93.0} & \refrow{89.5} & \refrow{90.5} \\
\refrow{\pih + oracle subgoals~\cite{robomme}} & \refrow{ground truth} & \refrow{83.9} & \refrow{93.3} & \refrow{95.2} & \refrow{64.0} & \refrow{84.1} \\
\refrow{\pih + oracle memory + our routines} & \refrow{ground truth} & \refrow{95.0} & \refrow{98.0} & \refrow{96.5} & \refrow{94.5} & \refrow{96.0} \\
\midrule
\pih~\cite{robomme} & none & 28.8 & 17.0 & 17.2 & 8.8 & 17.9 \\
\midrule
\multicolumn{7}{@{}l}{\emph{Memory inside the policy}} \\
FrameSamp+Modul~\cite{robomme} & sampled frames & 65.2 & 25.1 & 36.3 & 51.4 & 44.5 \\
FrameSamp+Modul\rep & sampled frames & 70.0 & 27.0 & 37.5 & 49.5 & 46.0 \\
EvoMem-VLA~\cite{evomem} & state-change tokens & \second{86.5} & 88.5 & 74.5 & 78.5 & 82.0 \\
SimpleMemVLA~\cite{simplememvla} & video context & \best{91.5} & \best{96.0} & \second{82.5} & \second{83.0} & \second{88.3} \\
\midrule
\multicolumn{7}{@{}l}{\emph{Memory managed by a VLM above the policy}} \\
GroundSG+QwenVL~\cite{robomme} & subgoal history & 38.0 & 39.3 & 31.6 & 21.9 & 32.7 \\
MemER~\cite{robomme} & keyframes + subgoals & 48.8 & 53.2 & 38.0 & 29.5 & 42.4 \\
PonderPounce~\cite{ponderpounce} & latent MLLM context & 74.7 & 62.8 & 72.2 & 33.7 & 60.8 \\
Ledger~\cite{ledger} & object ledger + frames & 66.8 & 86.7 & 60.7 & 43.0 & 64.3 \\
SimpleARM~\cite{simplearm} & typed state & 57.0 & 78.7 & 78.5 & 54.5 & 67.2 \\
\midrule
\oursrow \textbf{\method (ours)} & verified task state & \best{91.5} & \second{93.0} & \best{94.0} & \best{93.0} & \best{92.9} \\
\bottomrule
\end{tabular}
\vspace{-10pt}
\end{table*}

\paragraph{Baselines}
On \robomme we compare with the reactive \pih, with memories inside the policy (the benchmark's FrameSamp+Modul, EvoMem-VLA and SimpleMemVLA), and with memories managed by a VLM above the policy (the benchmark's GroundSG+QwenVL, MemER~\cite{memer}, PonderPounce, Ledger and SimpleARM). On \rmbench we compare with recent methods that report its M(1) tasks and run Mem-0 from its released checkpoint. On the real robot we compare with four baselines trained on the same demonstrations and evaluated in the same session: \pih, MemoryVLA~\cite{memoryvla}, MemER, and a \pih variant that also receives the subtask sentences it predicted earlier (\pih + subtask history).

\paragraph{Implementation}
To measure the memory rather than the policy, the executor is fixed on each benchmark. On \robomme the executor is the benchmark's released subgoal-conditioned \pih checkpoint, used as is; GroundSG+QwenVL, SimpleARM and the oracle use the same released policy. The pixel verifier is fitted on 4{,}640 rollouts of this executor in the training scenes. The subgoal writer is the VLM on eight tasks and a template on the other eight. On nine tasks a short agent-written motion routine can act on the verified target; it reads only camera images and joint states. The policy still issues at least 98\% of the executed action chunks on eleven tasks, where a routine is a rare fallback or only snaps the target. On StopCube, ButtonUnmaskSwap, PickHighlight and RouteStick a routine issues 77--93\% of the chunks, and on InsertPeg it performs the whole insertion, guided by a keypoint model (\cref{tab:supp_robomme}); on these five tasks the released policy reaches 56.9\% even with ground-truth subgoals, so execution limits it there, and the routine removes that limit from the measurement of memory. On \rmbench, for the same reason, one \pih is fine-tuned for all five tasks on 2{,}018 demonstrations with subgoal prompts and rendered cues, and subgoals come from templates. The memory layer has no trained component on \rmbench. On the real robot \pih is fine-tuned on 70 (two-bin) and 89 (bean scooping) teleoperated demonstrations, and the writer is a template.

\subsection{Benchmark results}
\label{sec:results}

\paragraph{Results on \robomme}
As shown in \cref{tab:robomme}, \method succeeds in 743 of 800 episodes (92.9\%), 4.6 points above the best published result, SimpleMemVLA (88.3\%). \method is the only method above 90\% in all four suites, also in a second full run (92.4\%, 739/800). With the same executor and routines, ground-truth memory from the simulator solves 768 episodes (96.0\%): \method is within 3.1 points of it. On the eleven tasks that the policy executes, \method reaches 93.3\% (513/550), against 90.2\% for SimpleMemVLA. Of the systems that share the released policy and differ only in how the task state is written, GroundSG+QwenVL reaches 44.1\% there and SimpleARM 81.3\%; ground-truth subgoals from the simulator give 96.4\%. With that policy executing all 16 tasks, they reach 32.7\% and 67.2\%, and \method 80.9\% (78.0\% in a second run), against 84.1\% for ground-truth subgoals.

\begin{table}[t]
\centering
\caption{\textbf{Results on the M(1) tasks of \rmbench} (success rate, \%). Published rows use 100 episodes per task (DIM-WAM and MemoryWAM as listed in~\cite{simplememvla}); \rep{} rows are run in our harness on 10 fixed test seeds per task (Mem-0\rep: mean of two runs). ``w/o memory'' removes the stored task state only; the last row uses 100 further seeds per task. \textbf{Bold}: best; \underline{underline}: second best.}
\label{tab:rmbench}
\footnotesize
\setlength{\tabcolsep}{2.4pt}
\begin{tabular}{@{}lcccccc@{}}
\toprule
Method & Obs. & Rearr. & PutB. & Swap & SwapT & Avg. \\
\midrule
\pih~\cite{rmbench} & 9 & 13 & 11 & 24 & 15 & 14.4 \\
Mem-0~\cite{rmbench} & 4 & 89 & 90 & 67 & 14 & 52.8 \\
Mem-0\rep & 0 & 90 & \best{100} & 65 & 10 & 53.0 \\
EventVLA (anchors)~\cite{eventvla} & 21 & 96 & 95 & \second{96} & 87 & 79.0 \\
DIM-WAM~\cite{dimwam} & 13 & \second{99} & \second{98} & \second{96} & \second{97} & 80.6 \\
EvoMem-VLA~\cite{evomem} & 28 & \best{100} & 92 & \best{100} & \best{100} & 84.0 \\
MemoryWAM~\cite{memorywam} & 27 & \best{100} & \best{100} & \best{100} & 94 & 84.2 \\
SimpleMemVLA~\cite{simplememvla} & \best{65} & \best{100} & \best{100} & \best{100} & 93 & \best{91.6} \\
\midrule
\method w/o memory\rep & 30 & 60 & 20 & 20 & 0 & 26.0 \\
\oursrow \textbf{\method (ours)}\rep & \second{60} & 80 & \best{100} & \best{100} & \best{100} & \second{88.0} \\
\oursrow \method, 100 further seeds & 65 & 84 & 100 & 89 & 99 & 87.4 \\
\bottomrule
\end{tabular}
\end{table}

\paragraph{Results on \rmbench}
On the five M(1) tasks (\cref{tab:rmbench}), \method succeeds in 44 of 50 episodes (88.0\%), against 53.0\% for the released Mem-0 on the same seeds, and is second only to SimpleMemVLA, with one shared, architecturally unchanged \pih. There it solves every episode of PutBack, Swap and SwapT, in which an initial layout must be recalled; without the task state these fall to 20\%, 20\% and 0\% (rollouts in \cref{fig:sim}). On 100 further seeds per task it solves 437 of 500 (87.4\%). These results answer Q1: \method leads on \robomme, on the full benchmark and on the tasks its policy executes, and is second on \rmbench.

\begin{figure*}[p]
    \centering
    \includegraphics[width=\linewidth]{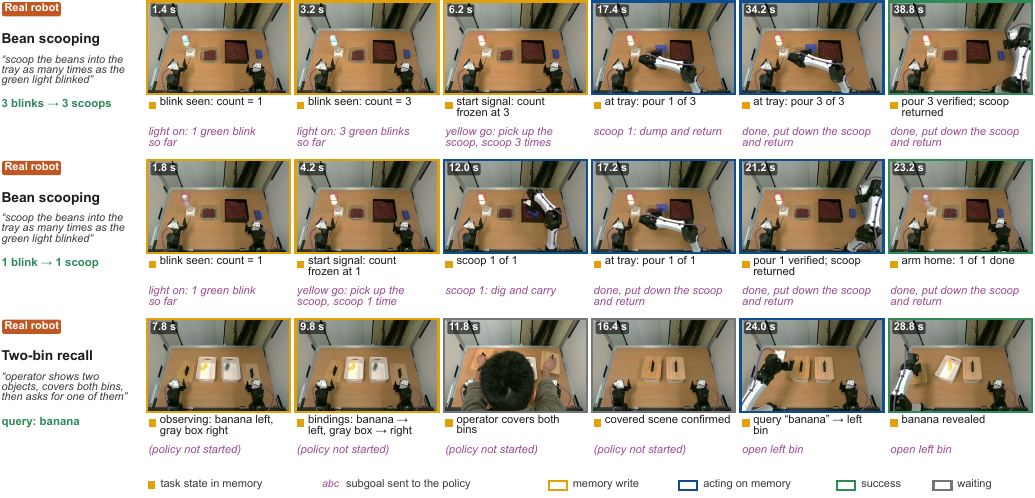}
    \caption{\textbf{Rollouts of \method on the real robot} (top camera). Rows 1--2: bean scooping after three blinks and after one blink (30/30 trials succeed, 10 per blink count). The blink counter writes the count, which is frozen at the start signal, and every later subgoal depends on it. Row 3: two-bin recall (30/30 trials). The bindings are written while the bins are open, and the query arrives after the operator has covered them. Under each frame, the first line is the task state held in memory and the purple line is the language subgoal sent to the policy. Border colors mark memory writes (gold), actions that depend on the memory (blue), waiting (gray) and success (green). The small pink box marks the lamp region read by the blink counter.}
    \label{fig:real}
\end{figure*}

\begin{figure*}[p]
    \centering
    \includegraphics[width=\linewidth]{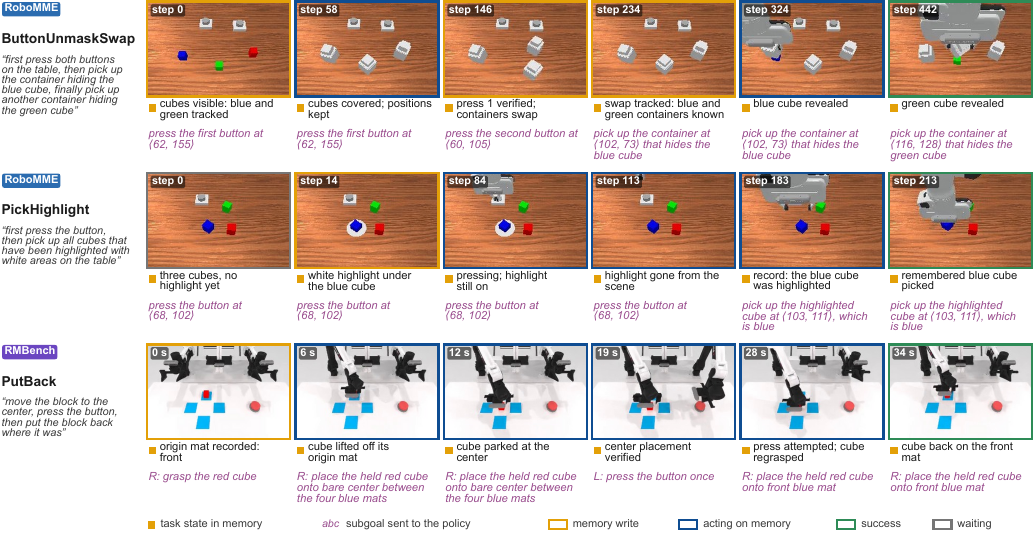}
    \caption{\textbf{Rollouts of \method in simulation.} Rows 1--2: \robomme (front camera); row 3: \rmbench. Captions and border colors as in \cref{fig:real}; R and L denote the right and left arm. Further rollouts are in the supplementary material.}
    \label{fig:sim}
\end{figure*}

\subsection{Real-robot experiments}
\label{sec:real}

\begin{figure}[t]
    \centering
    \includegraphics[width=\linewidth]{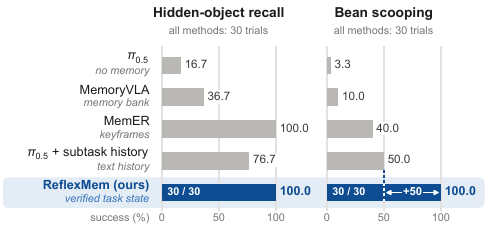}
    \caption{\textbf{Real-robot comparison.} All methods use the same demonstrations, the same robot and session, and 30 trials per task.}
    \label{fig:realbars}
\vspace{-10pt}
\end{figure}

In \textbf{two-bin recall}, a color-based detector writes the two bindings while the lids are open, and a template selects the lid to open once the scene is confirmed covered; all 30 consecutive trials succeed, with the query and the object positions varied. In \textbf{bean scooping}, the blink counter writes the count at camera rate and freezes it at the start signal, and each pour is committed from the measured wrist roll until the count is reached; all 30 trials succeed, 10 per blink count (\cref{fig:real}).

\Cref{fig:realbars} compares with the four baselines. On hidden-object recall, \method and MemER complete every trial; bean scooping separates the methods. There the strongest baselines reach 50.0\% (\pih + subtask history) and 40.0\% (MemER), and \method completes all 30 trials. This task pairs a cue of 0.3\,s flashes with near-identical scoop cycles; \method writes the blink count from every frame and advances the pour count only on a measured pour. This answers Q2: both paths carry over.

\subsection{Ablation study}
\label{sec:ablation}

\paragraph{The memory is necessary}
Removing the task state while keeping executor, prompts and seeds reduces the success rate on \rmbench from 88.0\% to 26.0\% (\cref{tab:rmbench}), below Mem-0 (53.0\%). On 320 \robomme test episodes, with the motion routines active, success falls from 93.4\% to 4.1\%.

\paragraph{Verification is necessary}
Here the released policy executes every subgoal of all 16 tasks. Without verification, committing each subgoal after the number of actions it typically takes in training solves 376 of the 800 episodes, and committing after the first action chunk solves 94; the pixel verifier solves 647 (80.9\%) and the simulator's signal 651; in the full system they solve 743 and 745.

\paragraph{Reading on state change}
Calling the writer at every chunk, with the subgoal kept unless its meaning or target changes, solves 290 of 400 episodes of the VLM-writer tasks (state-triggered: 365) with 29.2 instead of 4.6 calls per episode.

\paragraph{A new task is configured, not trained}
A new coding agent receives only a task's instruction, ten training demonstrations and the program library, and may validate at most eight revisions on held-out training episodes; no model is trained. The library holds none of our task logic: the agent writes its own, e.g.\ the route reading of RouteStick. On the 15 \robomme tasks other than InsertPeg, its configurations solve 708 of 750 test episodes (94.4\%), against 699 for ours, in 2.3 rounds and 32 minutes each (\cref{sec:supp_agent}).

\paragraph{The write path under image changes}
On \robomme, shifting the images of programs and verifier by 4 pixels, scaling their brightness by 0.8 or 1.2, or adding noise leaves 440, 442, 436 and 429 of 480 episodes solved, against 448 (\cref{sec:supp_cost}).

\begin{figure}[t]
    \centering
    \includegraphics[width=\linewidth]{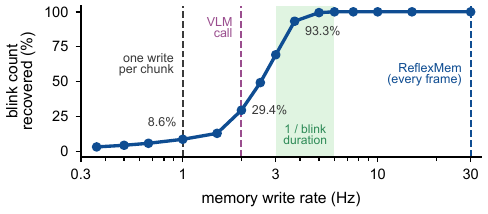}
    \caption{\textbf{Write rate vs.\ remembered count.} Share of the 89 bean-scooping demonstrations whose blink count a per-frame detector recovers at the given rate.}
    \label{fig:rate}
\vspace{-10pt}
\end{figure}

\paragraph{The write rate decides what can be remembered}
We replay the 89 bean-scooping demonstrations through the blink counter on every $k$-th frame (\cref{fig:rate}), an upper bound for any writer that sees one frame per write. The count is recovered down to 6\,Hz and falls below it: 29\% at one write per VLM call (0.5\,s) and 8.6\% at one per 30-action chunk (1\,Hz). In closed loop on \robomme, one program pass per action chunk instead of every 2--8 steps reduces success from 448 to 381 of 480 episodes, and one per two chunks to 264. This answers Q3: the task state, verification, fast writing and state-triggered reading each matter.

\subsection{Efficiency analysis}
\label{sec:efficiency}

\begin{table}[t]
\centering
\caption{\textbf{Run-time cost} on one A100 (medians). \emph{Decision}: observation to next action chunk with unchanged task state. \emph{Mean}: per decision over 80 \robomme episodes, with VLM calls. $^{\ddagger}$RTX~4090.}
\label{tab:cost}
\footnotesize
\setlength{\tabcolsep}{2.4pt}
\begin{tabular}{@{}lccc@{}}
\toprule
System & Write & Decision & Mean \\
\midrule
\pih (no memory) & -- & 75\,ms & 87\,ms \\
FrameSamp+Modul & 50\,ms / chunk & 144\,ms & 437\,ms \\
VLM writer, one call & 0.52\,s & -- & -- \\
\midrule
\oursrow \method, \robomme & 9.3\,ms / obs.\ + 7\,ms / chunk & 82\,ms & 195\,ms \\
\oursrow \method, real robot$^{\ddagger}$ & 0.16\,ms / frame & 90\,ms & -- \\
\bottomrule
\end{tabular}
\end{table}

A decision with unchanged task state costs 82\,ms (\cref{tab:cost}), 7\,ms more than the policy alone. \method calls no VLM on the template tasks, on \rmbench, or on the real robot: an explicit, verified task state can be read by a template. Counting every writer call, \method averages 195\,ms per decision (FrameSamp+Modul: 437\,ms; SimpleMemVLA: 0.68\,s on an H100~\cite{simplememvla}). This answers Q4: a program pass costs 9.3\,ms on \robomme and 0.16\,ms on the robot, up to three orders of magnitude below a VLM call.

%% file: sec/5_conclusion.tex
\section{Conclusion}
\label{sec:conclusion}

We presented \method, a memory layer for agentic VLA systems that writes a verified task state at the rate of the world and reads it on change. With an architecturally unchanged \pih, it completes 60 of 60 real-robot trials, achieves 92.9\% on \robomme and 88.0\% on \rmbench M(1), and writes memory up to three orders of magnitude cheaper than a VLM call. \textbf{Limitation.} Configurations are written per task, not yet at run time.

%% file: sec/6_llm_statement.tex
\section*{LLM Usage Statement}
Language models (Claude, Anthropic) were used in assistance with implementation, debugging, manuscript editing and polishing. The authors checked every reported number against the result files, verified every citation against the cited paper, and are responsible for all scientific claims and the final submission.

%% file: sec/X_suppl.tex
\clearpage
\maketitlesupplementary
\appendix
\makeatletter\setlength{\@fptop}{0pt}\setlength{\@dblfptop}{0pt}\setlength{\@fpsep}{16pt}\setlength{\@dblfpsep}{16pt}\makeatother 

\noindent This document provides per-task results and configurations (\cref{sec:supp_tasks}), details of the perception programs and the task state (\cref{sec:supp_programs}), the pixel verifier (\cref{sec:supp_verifier}), the subgoal writer (\cref{sec:supp_exec}), the configuration by a fresh agent (\cref{sec:supp_agent}), the real-robot protocol (\cref{sec:supp_real}), the measurement protocol of the efficiency analysis (\cref{sec:supp_cost}), and additional rollouts (\cref{sec:supp_rollouts}). Citation numbers refer to the reference list of the main paper.

\section{Per-task results and configuration}
\label{sec:supp_tasks}

\begin{table*}[t]
\centering
\caption{\textbf{Per-task results on \robomme} (success rate, \%; ``unseen seeds'': successes out of 20 episodes with unseen scene seeds). Task abbreviations: BinFill (BF), PickXtimes (PX), SwingXtimes (SX), StopCube (SC), VideoUnmask (VU), ButtonUnmask (BU), VideoUnmaskSwap (VUS), ButtonUnmaskSwap (BUS), PickHighlight (PH), VideoRepick (VR), VideoPlaceButton (VPB), VideoPlaceOrder (VPO), MoveCube (MC), InsertPeg (IP), PatternLock (PL), RouteStick (RS). The rows above \method are the systems that share the benchmark's released subgoal-conditioned policy, as reported by their authors; the oracle-memory row is our run. The three bottom rows give the configuration of \method per task: writer type (V: VLM, T: template), the routine that is added to the released policy (--: none), and the share of executed action chunks that a routine issues in the 800 test episodes.}
\label{tab:supp_robomme}
\footnotesize
\setlength{\tabcolsep}{3.0pt}
\begin{tabular}{@{}l cccc cccc cccc cccc c@{}}
\toprule
 & \multicolumn{4}{c}{Counting} & \multicolumn{4}{c}{Permanence} & \multicolumn{4}{c}{Reference} & \multicolumn{4}{c}{Imitation} & \\
\cmidrule(lr){2-5}\cmidrule(lr){6-9}\cmidrule(lr){10-13}\cmidrule(lr){14-17}
Method & BF & PX & SX & SC & VU & BU & VUS & BUS & PH & VR & VPB & VPO & MC & IP & PL & RS & Avg. \\
\midrule
\refrow{\pih + oracle subgoals~\cite{robomme}} & \refrow{85.8} & \refrow{100} & \refrow{100} & \refrow{49.7} & \refrow{98.8} & \refrow{95.0} & \refrow{99.2} & \refrow{80.2} & \refrow{83.3} & \refrow{97.3} & \refrow{100} & \refrow{100} & \refrow{87.8} & \refrow{15.6} & \refrow{97.0} & \refrow{55.6} & \refrow{84.1} \\
\refrow{\pih + oracle memory + our routines} & \refrow{82} & \refrow{100} & \refrow{98} & \refrow{100} & \refrow{98} & \refrow{100} & \refrow{100} & \refrow{94} & \refrow{98} & \refrow{92} & \refrow{100} & \refrow{96} & \refrow{94} & \refrow{88} & \refrow{98} & \refrow{98} & \refrow{96.0} \\
GroundSG+QwenVL~\cite{robomme} & 52.0 & 92.7 & 7.3 & 0.0 & 88.7 & 24.0 & 30.7 & 14.0 & 15.1 & 25.3 & 54.0 & 31.8 & 71.6 & 3.3 & 6.7 & 6.0 & 32.7 \\
SimpleARM~\cite{simplearm} & 60.0 & 91.3 & 76.7 & 0.0 & 96.0 & 82.0 & 80.0 & 56.7 & 66.0 & 78.0 & 82.0 & 88.0 & 67.3 & 4.0 & 93.3 & 53.3 & 67.2 \\
\midrule
\oursrow \textbf{\method} & 90 & 100 & 88 & 88 & 100 & 90 & 94 & 88 & 98 & 90 & 94 & 94 & 86 & 88 & 100 & 98 & 92.9 \\
\method, second run & 90 & 100 & 96 & 88 & 98 & 90 & 92 & 86 & 94 & 90 & 94 & 94 & 82 & 88 & 98 & 98 & 92.4 \\
\method, unseen seeds (/20) & 15 & 19 & 17 & 19 & 20 & 18 & 20 & 19 & 18 & 20 & 17 & 18 & 16 & 18 & 20 & 20 & {\scriptsize 294/320} \\
\midrule
Writer & V & V & V & T & T & V & T & V & V & V & T & T & V & T & T & T & \\
Added routine & a & -- & -- & b & -- & -- & a & c & c & d & -- & -- & a & e & -- & c & \\
Routine chunks (\%) & 2 & 0 & 0 & 81 & 0 & 0 & 1 & 89 & 77 & 0 & 0 & 0 & 0 & 100 & 0 & 93 & \\
\bottomrule
\multicolumn{18}{@{}l}{\scriptsize a: one programmed pick/place/press on the verified target if the policy has not completed the subgoal within its action budget; b: timing rule for the press and a hover gate;} \\
\multicolumn{18}{@{}l}{\scriptsize c: programmed primitive on the verified target; d: target coordinates snapped to the tracked object; e: keypoint model and programmed insertion (no policy calls).} \\
\multicolumn{18}{@{}l}{\scriptsize All routines are written by the coding agent.}
\end{tabular}
\end{table*}

\paragraph{\robomme}
\Cref{tab:supp_robomme} lists per-task success rates, a second run, the results on unseen scene seeds, and the configuration of each task. The second run repeats the 800 test episodes with the unchanged system: 739 succeed (suites: 93.5, 91.5, 93.0 and 91.5\%), and 730 episodes succeed in both runs. The policy issues at least 98\% of the executed chunks on eleven tasks (all but SC, BUS, PH, IP and RS), which are those of the comparison in the main paper: 513 of 550 episodes succeed (512 in the second run). On these tasks the mean of the oracle row is 96.4\%, that of GroundSG+QwenVL 44.1\%, and that of SimpleARM 81.3\%, and SimpleMemVLA reports 90.2\%~\cite{simplememvla}; the three rows use the benchmark's released subgoal-conditioned \pih policy (the benchmark's own rows average its last three checkpoints). On the seven tasks without any routine (PX, SX, VU, BU, VPB, VPO, PL), 333 of 350 episodes succeed. In the oracle-memory row, the subgoal and its target, the press time of StopCube, the grasp facts of InsertPeg and the completion of each subgoal come from the simulator, and the executor and the routines are those of \method: 768 of 800 episodes succeed. With only the verifier's decision replaced by the simulator's completion signal, 745 succeed. No routine reads simulator state: each uses the camera images and the joint state; a guard that blocks reads of simulator objects during decisions recorded none in the 800 test episodes. The episodes with unseen seeds are fresh scenes for which the benchmark's demonstration generator produced a valid demonstration. All learned components of \method are fitted on the training split only; the motion routine of each task is part of its configuration and is listed in \cref{tab:supp_robomme}. The memory-removed control of the main paper (episodes 0--19 of every task) removes the commit log, the stored facts and the cursor and keeps the executor, the verifier and the routines: 13 of 320 episodes succeed, against 299 for the full system.

\begin{table}[t]
\centering
\caption{\textbf{\rmbench M(1): additional runs in our harness} (successes out of 10 fixed seeds per task).}
\label{tab:supp_rmbench}
\footnotesize
\setlength{\tabcolsep}{2.6pt}
\resizebox{\dimexpr\linewidth-2\tabcolsep}{!}{%
\begin{tabular}{@{}lcccccc@{}}
\toprule
Run & Obs. & Rearr. & PutB. & Swap & SwapT & Total \\
\midrule
Mem-0, released (2 runs) & 0 & 9 & 10 & 6.5 & 1 & 26.5 \\
\midrule
w/o memory & 3 & 6 & 2 & 2 & 0 & 13 \\
w/o memory, current-frame cue & 3 & 6 & 2 & 3 & 0 & 14 \\
\oursrow \method & 6 & 8 & 10 & 10 & 10 & 44 \\
\bottomrule
\end{tabular}}
\end{table}

\paragraph{\rmbench}
\Cref{tab:supp_rmbench} gives the raw counts behind the main table. On 100 further seeds per task, seeds that the benchmark's expert solves from ranges never used before, \method succeeds in 65, 84, 100, 89 and 99 episodes (Observe\&PickUp, Rearrange, PutBack, Swap, SwapT), 437 of 500. The two memory-removed rows use the executor and seeds of the full system; the write path re-perceives from the current frame at every decision, and in the second row the rendered cue is recomputed from the current frame. The \rmbench executor is one \pih, fine-tuned from the public base checkpoint with full fine-tuning for 24{,}000 updates (batch size 32, EMA 0.99, cosine learning rate with a peak of $2.5{\times}10^{-5}$) on 2{,}018 demonstrations of the five tasks, all from training seeds. The \pih and Mem-0 rows of the main table are taken directly from the \rmbench paper. The memory layer contains no trained model: subgoals follow a stage library mined from training demonstrations; placements and releases are confirmed from measured joint states.

\section{Perception programs and task state}
\label{sec:supp_programs}

Every program maps an image region or the joint state to a signal or a fact (main paper, Sec.~3.2). The main programs are the following.
\begin{itemize}
    \item \emph{Blink counter}: mean green and yellow color excess in a region around the lamp, relative to a baseline from the first frames, with on and off thresholds for a blink; the start signal fires on sustained yellow excess and freezes the count.
    \item \emph{Blob tracker}: connected components per color on the table region after subtracting an image of the empty table, with centroid and pixel count; a \emph{containment test} reports whether a tracked blob lies inside a container footprint.
    \item \emph{Histogram matcher}: color-histogram intersection between an exemplar (\eg, the object on the shelf in Observe\&PickUp) and each table blob, used to re-identify a remembered object.
    \item \emph{Bin-content detector}: color rules for the objects inside each bin region; a binding is committed after consistent observations, and the scene must be confirmed covered before a query is accepted.
    \item \emph{Pour detector}: the wrist-roll excursion relative to its value at the start of the current scoop must exceed an upper threshold and return below a lower one.
    \item \emph{Phase detector}: distance of the arm to recorded reference poses, with a speed-based settle test, and a \emph{release detector} on the gripper opening.
\end{itemize}
On the real robot the programs run on every camera frame (30\,Hz); on \robomme they run on sampled observations between chunk boundaries (every 2 to 8 environment steps), and on \rmbench at every decision (every 10 actions). The task state holds counters, bindings, a cursor into the instruction and a log of commits; \cref{fig:supp_robomme_a,fig:supp_robomme_b,fig:supp_robomme_c} print the state next to each frame.

\section{Pixel verifier}
\label{sec:supp_verifier}

\begin{table}[t]
\centering
\caption{\textbf{Verification ablation on \robomme}: successes on the 800 test episodes with the released policy executing every subgoal of all 16 tasks.}
\label{tab:ablation}
\footnotesize
\setlength{\tabcolsep}{4pt}
\begin{tabularx}{\dimexpr\linewidth-2\tabcolsep}{@{}Xc@{}}
\toprule
Completion signal & Success / 800 \\
\midrule
\refrow{Simulator (oracle)} & \refrow{651} \\
None (commit on command) & 94 \\
Timer & 376 \\
\oursrow Pixel verifier (ours) & \textbf{647} \\
\bottomrule
\end{tabularx}
\end{table}

\paragraph{Without verification}
In the configuration of \cref{tab:ablation}, we discard the verifier's decision in two ways. \emph{Commit on command} commits a subgoal as soon as its first action chunk has been executed: 94 of the 800 episodes succeed (38 and 27 of them on VideoUnmask and VideoUnmaskSwap, none on nine tasks), and the verifier would have withheld 18{,}886 of the 18{,}984 commits. The \emph{timer} commits a subgoal after the median number of actions that subgoals of the same task and verb take to complete in the training rollouts, rounded up to a multiple of 16. On the 800 episodes the timer solves 376 and the pixel verifier 647; 366 episodes are solved by both, 10 by the timer only and 281 by the verifier only. The largest differences are on PickXtimes (0 against 49), VideoRepick (0 against 45), SwingXtimes (11 against 47) and PickHighlight (10 against 40).

\paragraph{Architecture}
A ResNet-18 trunk (ImageNet initialization) encodes $160{\times}160$ images with five channels: RGB and two Gaussian target heatmaps ($\sigma \approx 6$ pixels) on the front view; wrist images receive empty heatmaps. Eight image tokens (the start of the current subgoal and the three most recent chunk boundaries, for two cameras) carry learned camera and time embeddings and are pooled with a few context tokens by a two-layer transformer (width 256, four heads). One head predicts completion; an auxiliary head predicts the color of the held object. The model has 12.4M parameters.

\paragraph{Data and training}
We roll out the released executor on the training scenes (4{,}640 rollouts, 290 per task) and label every chunk boundary with the simulator's completion signal, which yields 91{,}348 checks (12{,}227 positive). Scenes are split 966/244 for training and validation. We train for 12 epochs with AdamW (learning rate $10^{-4}$ for the trunk and $3{\times}10^{-4}$ for the heads), binary cross-entropy plus 0.3 times the cross-entropy of the auxiliary head, task-balanced sampling with one positive per four negatives, in bf16; one model takes 0.4 GPU-hours. On held-out scenes (18{,}149 checks from 244 scenes) it recalls 95.2\% of the completed subgoals at a threshold that keeps premature commits at or below 2\% of subgoal segments; at the operating point the precision is 97.3\%. A forward pass takes 2.5\,ms at batch size 1 on an L40S, and 7\,ms in the loop on the A100 of the efficiency analysis, where two evaluation streams share the GPU. \Cref{tab:ablation} gives the closed-loop ablation of the completion signal discussed in the main paper.

\section{Subgoal writer}
\label{sec:supp_exec}

\paragraph{VLM writer}
The VLM writer is Qwen3-VL-4B with a low-rank adapter trained for one epoch on the training split (instruction, facts, log of commits and current frame as input; the next grounded subgoal as output). It is called after a commit or when the current subgoal exceeds its action budget; on the eight tasks that use it this amounts to 4.6 calls per episode on average, against 26.9 chunk boundaries.

\paragraph{Reading at every chunk}
In this control the writer is also called after every verifier check that does not commit, with the same request as at a state change (current frame, goal, verified history, facts). Its output replaces the current subgoal only if the verb or the object changes or a target moves by more than 12 pixels, a rule fixed before the run; commits and motion routines are unchanged. On the 400 episodes of the eight VLM-writer tasks, 290 succeed, against 365 for state-triggered reading (BinFill 14 against 45, ButtonUnmask 41/45, ButtonUnmaskSwap 43/44, MoveCube 40/43, PickHighlight 47/49, PickXtimes 45/50, SwingXtimes 46/44, VideoRepick 14/45), with 29.2 instead of 4.6 writer calls per episode. About a quarter of the additional calls replace the subgoal, more than half of them with another verb: called before the current subgoal is verified, the writer alternates between consecutive steps, such as picking up and putting down in VideoRepick.

\section{Configuration by a fresh agent}
\label{sec:supp_agent}

\begin{table}[t]
\centering
\caption{\textbf{Configurations written by a fresh coding agent} on the 15 \robomme tasks other than InsertPeg: successes on the 50 test episodes (one run of the frozen configuration) next to those of the configurations of \cref{tab:supp_robomme} on the same episodes; validation rounds used (of eight allowed).}
\label{tab:supp_agent}
\footnotesize
\setlength{\tabcolsep}{3.2pt}
\begin{tabular}{@{}lccc@{}}
\toprule
Task & Agent & Ours & Rounds \\
\midrule
BinFill & 49 & 45 & 2 \\
PickXtimes & 50 & 50 & 2 \\
SwingXtimes & 47 & 44 & 1 \\
StopCube & 50 & 44 & 4 \\
VideoUnmask & 49 & 50 & 3 \\
ButtonUnmask & 50 & 45 & 3 \\
VideoUnmaskSwap & 50 & 47 & 1 \\
ButtonUnmaskSwap & 48 & 44 & 4 \\
PickHighlight & 45 & 49 & 2 \\
VideoRepick & 48 & 45 & 1 \\
VideoPlaceButton & 34 & 47 & 2 \\
VideoPlaceOrder & 44 & 47 & 4 \\
MoveCube & 44 & 43 & 2 \\
PatternLock & 50 & 50 & 1 \\
RouteStick & 50 & 49 & 2 \\
\midrule
Total (750) & 708 & 699 & 34 \\
\bottomrule
\end{tabular}
\end{table}

For each of the 15 tasks, a new instance of the coding agent (Claude Opus 5.5) receives the task instruction, ten training demonstrations with the benchmark's annotations, and the task-independent program library with its documentation: perception primitives, camera geometry, the task-state store, the interfaces of the subgoal writer and the verifier, and motion primitives. The library does not contain the task logic of our configurations: each agent writes its own, for example the route reading of RouteStick, the pattern order of PatternLock, the press timing of StopCube and the tracking through swaps of VideoRepick and VideoUnmaskSwap. None of our configurations and no test seed is in this material, and the agent reaches the cluster only through a broker restricted to its task. It writes one program for the task and may submit at most eight revisions for validation on ten held-out training episodes. A validation report contains camera frames and the system's own logs, and a revision replaces the incumbent only if it scores higher. The final configuration is frozen and run once on the 50 test episodes.

\Cref{tab:supp_agent} lists the results. The agent's configurations solve 708 of 750 test episodes (94.4\%), against 699 (93.2\%) for ours; fourteen reach 10 of 10 on validation and VideoPlaceButton 8 of 10. No agent uses more than four rounds, and a task takes 32 minutes of wall-clock time on average.

\section{Real-robot protocol}
\label{sec:supp_real}

\paragraph{Platform}
All hardware experiments use a dual-arm YAM robot; each arm has six joints and a gripper (seven actuated DoF) and is driven at 30\,Hz with joint-position targets. We teleoperate demonstrations with GELLO. Three cameras (one fixed top view and one on each wrist) deliver images that are letterboxed to $224{\times}224$ (from $640{\times}480$ in our runs). Every policy predicts 30-action chunks; the policy of \method runs on a workstation with one RTX~4090 (about 0.09\,s per inference); the perception programs and the task state run on the CPU of the same machine.

\paragraph{Tasks and training}
In two-bin recall the operator's query gives the object and never the side, so the lid can only be chosen from what was seen while the bins were open. The policy is fine-tuned for 2{,}000 steps on 70 demonstrations. In bean scooping the lamp flashes one, two or three times, which fixes how many scoops are required; the number never appears in the instruction, and the scene after the start signal looks the same for every count. The policy is fine-tuned for 6{,}000 steps on 89 demonstrations.

\begin{figure*}[t]
    \centering
    \includegraphics[width=0.9\linewidth]{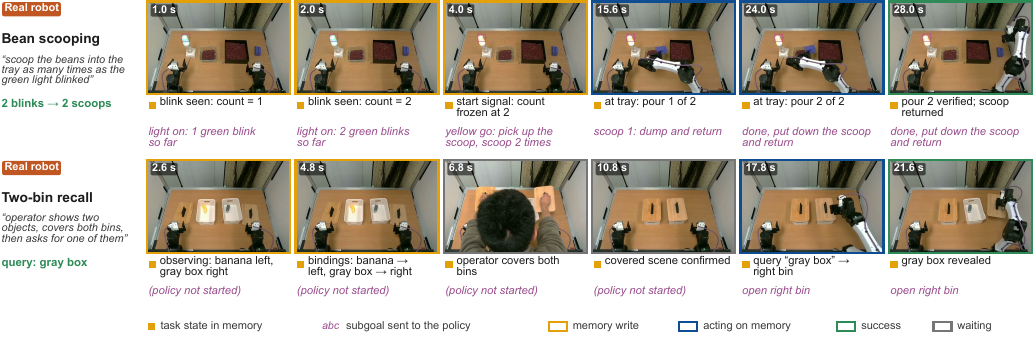}
    \caption{\textbf{Further real-robot trials}: bean scooping after two blinks, and two-bin recall with the other query.}
    \label{fig:supp_real}
\end{figure*}

\paragraph{Evaluation}
A trial counts only if the robot carries out the whole instruction; partial progress earns no credit. Two-bin recall is evaluated in 30 consecutive trials with a fresh process per trial, with the queried object varied and the object positions exchanged before some trials. Bean scooping is evaluated in 30 trials, 10 for each number of blinks. \Cref{fig:supp_real} shows the remaining trial types.

\begin{table}[t]
\centering
\caption{\textbf{Real-robot comparison}: successes / trials behind the figure of the main paper.}
\label{tab:supp_real}
\footnotesize
\setlength{\tabcolsep}{5pt}
\begin{tabular}{@{}lcc@{}}
\toprule
Method & Hidden-object recall & Bean scooping \\
\midrule
\pih & 5 / 30 & 1 / 30 \\
MemoryVLA & 11 / 30 & 3 / 30 \\
MemER & 30 / 30 & 12 / 30 \\
\pih + subtask history & 23 / 30 & 15 / 30 \\
\oursrow \textbf{\method (ours)} & 30 / 30 & 30 / 30 \\
\bottomrule
\end{tabular}
\end{table}

\paragraph{Baselines}
\Cref{tab:supp_real} lists the counts behind the real-robot comparison. All methods are evaluated in the same session on the same robot. \pih, MemoryVLA, MemER and \pih + subtask history (a \pih variant that also receives the subtask sentences it predicted earlier) are each trained per task on the same teleoperated demonstrations as \method (70 for hidden-object recall, 89 for bean scooping), executed with the same 30-action chunks, and scored over 30 trials per task with the same success criterion. Hidden-object recall uses two lidded bins for every method.

\section{Efficiency measurements}
\label{sec:supp_cost}

\begin{table}[t]
\centering
\caption{\textbf{Latency measurements} behind the efficiency table of the main paper (one A100-80GB, batch size 1, bf16, after warm-up; perception programs on a CPU).}
\label{tab:supp_cost}
\footnotesize
\setlength{\tabcolsep}{3.5pt}
\begin{tabular}{@{}lccc@{}}
\toprule
Component & Median & p10 / p90 & $N$ \\
\midrule
Released \pih policy, one chunk & 74.6\,ms & 74.2 / 75.8 & 40 \\
FrameSamp+Modul, encode 16 frames & 50.3\,ms & 48.0 / 54.1 & 40 \\
FrameSamp+Modul, one chunk & 93.6\,ms & 93.2 / 95.3 & 40 \\
VLM writer, one call & 0.52\,s & 0.29 / 0.73 & 400 \\
Pixel verifier, in loop & 7.1\,ms & 6.5 / 10.4 & 11{,}230 \\
Programs, \robomme, per observation & 9.3\,ms & 7.9 / 11.6 & 6{,}020 \\
Programs, bean scooping, per frame & 0.16\,ms & -- & 2{,}366 \\
\bottomrule
\end{tabular}
\end{table}

\Cref{tab:supp_cost} lists the measurements. All models are served from their released checkpoints with the official serving code and queried through the official client. FrameSamp+Modul follows the benchmark's evaluator: before each chunk the 16 newly executed frames are added to its history buffer and encoded. The VLM writer replays 400 requests recorded during our \robomme evaluation, each with one image, at a median of 0.52\,s per call. Over 80 \robomme episodes (five per task, one stream on an otherwise idle A100) and counting every writer call, the mean time per decision is 195\,ms for \method, 87\,ms for \pih alone and 437\,ms for FrameSamp+Modul. GroundSG+QwenVL, run as released in the benchmark's evaluator, takes 10.1\,s per decision (22.2 VLM calls per episode at a median of 3.28\,s). Verifier and program timings on \robomme are taken from the logs of the test episodes, and the bean-scooping program timing from a replay of recorded demonstrations.

\paragraph{Image perturbations}
Every image handed to the perception programs and to the pixel verifier is perturbed, and the configuration, the policy and the subgoal writer are unchanged; the policy and the writer receive the original image. The four perturbations are a shift by 4 pixels to the right and down with edge padding, brightness scaled by 0.8 and by 1.2, and Gaussian noise with $\sigma{=}5/255$ drawn per frame. On episodes 0--29 of all 16 tasks, 440, 442, 436 and 429 of 480 episodes succeed, against 448 without perturbation.

\paragraph{Write rate in closed loop}
On \robomme the perception programs of the final system run every five environment steps on the live stream (every two for the motion program of StopCube) and on every eighth frame of the conditioning video. In this control they run once per $S$ environment steps or video frames, and nothing else changes. \Cref{tab:supp_rate} reports episodes 0--29 of every task. With one pass per executed chunk ($S{=}16$), 67 of the 448 solved episodes are lost: the crossing counter of StopCube cannot follow the moving cube, and the tracks through the swaps of VideoRepick, VideoUnmaskSwap and ButtonUnmaskSwap and the order of the demonstrated events of VideoPlaceButton and VideoPlaceOrder break. With one pass per two chunks, 184 are lost, and with one per four chunks, 229.

\begin{table}[t]
\centering
\caption{\textbf{Write rate in closed loop on \robomme}: successes on episodes 0--29 with one perception pass per $S$ environment steps or video frames. The four tasks in the last rows have no observation stream to thin and are counted from the final system.}
\label{tab:supp_rate}
\footnotesize
\setlength{\tabcolsep}{4pt}
\begin{tabular}{@{}lcccc@{}}
\toprule
Task & Final & $S{=}16$ & $S{=}32$ & $S{=}64$ \\
\midrule
StopCube & 28 & 0 & 0 & 0 \\
ButtonUnmask & 26 & 25 & 11 & 13 \\
ButtonUnmaskSwap & 27 & 23 & 7 & 9 \\
PickHighlight & 30 & 29 & 28 & 7 \\
VideoUnmask & 30 & 30 & 30 & 29 \\
VideoUnmaskSwap & 28 & 23 & 6 & 6 \\
VideoRepick & 26 & 13 & 7 & 9 \\
VideoPlaceButton & 29 & 23 & 10 & 8 \\
VideoPlaceOrder & 28 & 22 & 7 & 7 \\
MoveCube & 27 & 27 & 18 & 7 \\
PatternLock & 30 & 29 & 30 & 14 \\
RouteStick & 29 & 27 & 0 & 0 \\
\midrule
BinFill, PickXtimes, SwingXtimes & 84 & 84 & 84 & 84 \\
InsertPeg & 26 & 26 & 26 & 26 \\
\midrule
All 16 tasks (480) & 448 & 381 & 264 & 219 \\
\bottomrule
\end{tabular}
\end{table}

\paragraph{Write-rate experiment}
For each of the 89 bean-scooping demonstrations we compute the lamp signals on all frames up to the start signal. A per-frame detector with the thresholds and the baseline of the deployed blink counter then sees only the frames $\varphi, \varphi{+}k, \varphi{+}2k, \dots$, for every stride $k \in \{1, 2, 3, 4, 5, 6, 8, 10, 12, 15, 20, 30, 45, 60, 83\}$ and every phase $\varphi < k$. A blink is counted when at least one sampled frame is lit and consecutive blinks are separated by a sampled dark frame, which is the most that any writer can detect frame by frame at that rate. The count is recovered in all demonstrations and phases for $k \le 5$, and in 99.4, 93.3, 69.2, 49.3, 29.4, 12.9, 8.6, 5.7, 4.3 and 3.1\% of the demonstration-phase pairs for the remaining strides. For $k \ge 20$ only demonstrations with a single blink (27 of 89) are ever counted correctly, and all errors are under-counts. Blinks last 5 to 10 frames (median 9) and are separated by 9 to 23 dark frames.

\section{Additional rollouts}
\label{sec:supp_rollouts}

\begin{figure*}[p]
    \centering
    \includegraphics[width=\linewidth]{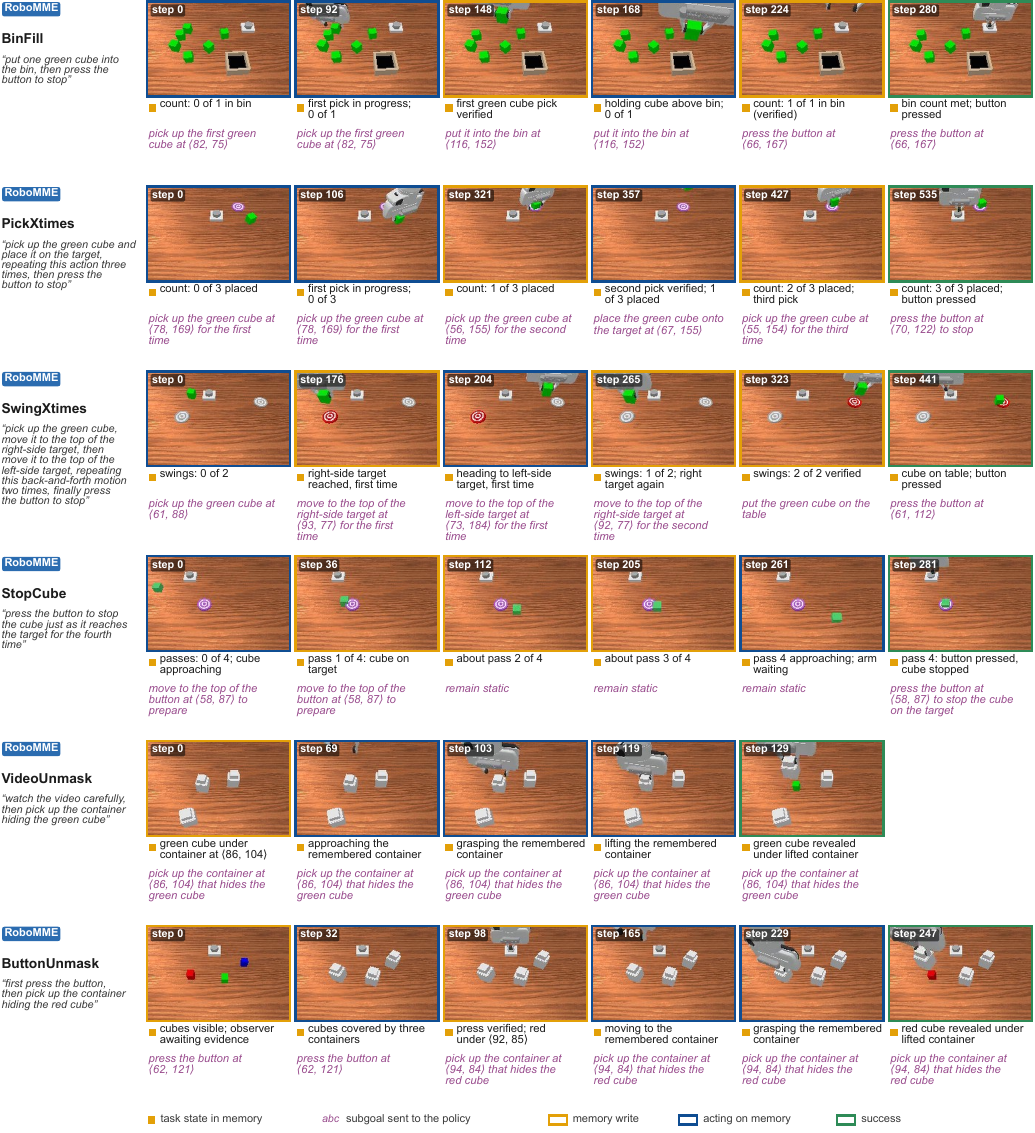}
    \caption{\textbf{\robomme rollouts (1/3)}: Counting suite and two Permanence tasks.}
    \label{fig:supp_robomme_a}
\end{figure*}

\begin{figure*}[p]
    \centering
    \includegraphics[width=\linewidth]{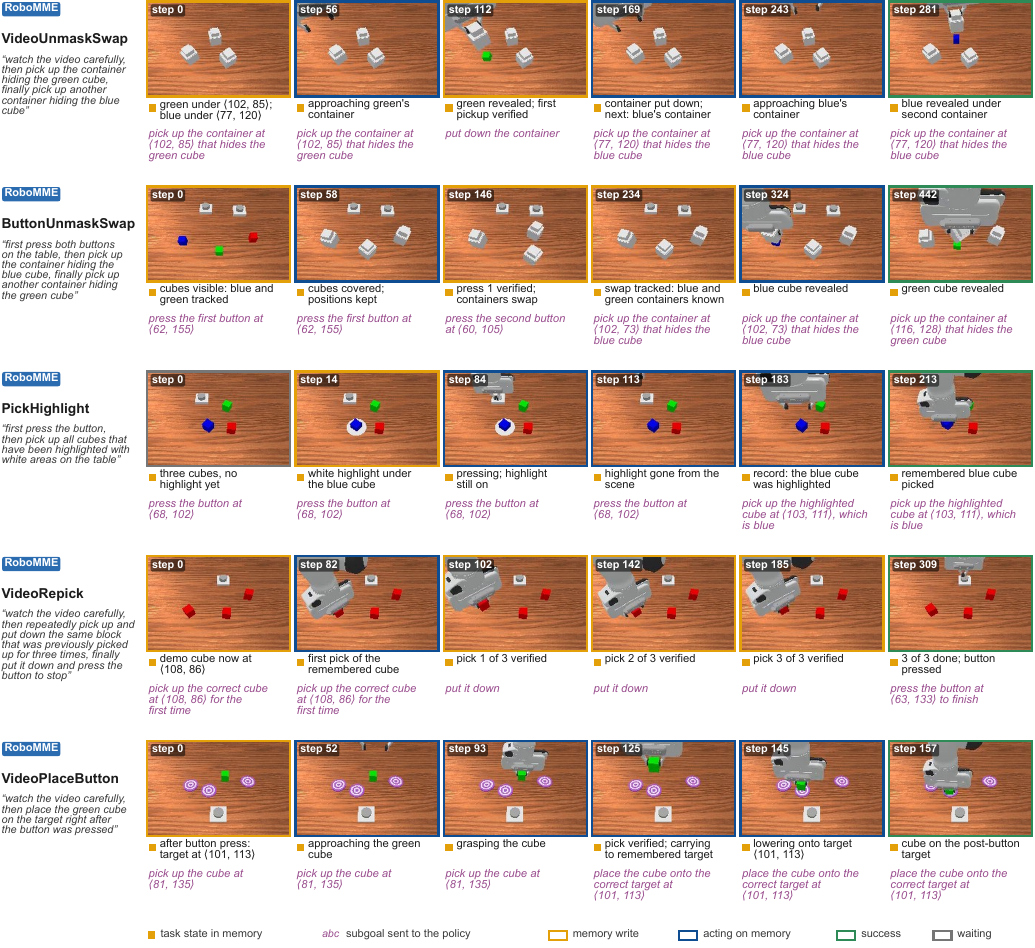}
    \caption{\textbf{\robomme rollouts (2/3)}: Permanence and Reference suites.}
    \label{fig:supp_robomme_b}
\end{figure*}

\begin{figure*}[p]
    \centering
    \includegraphics[width=\linewidth]{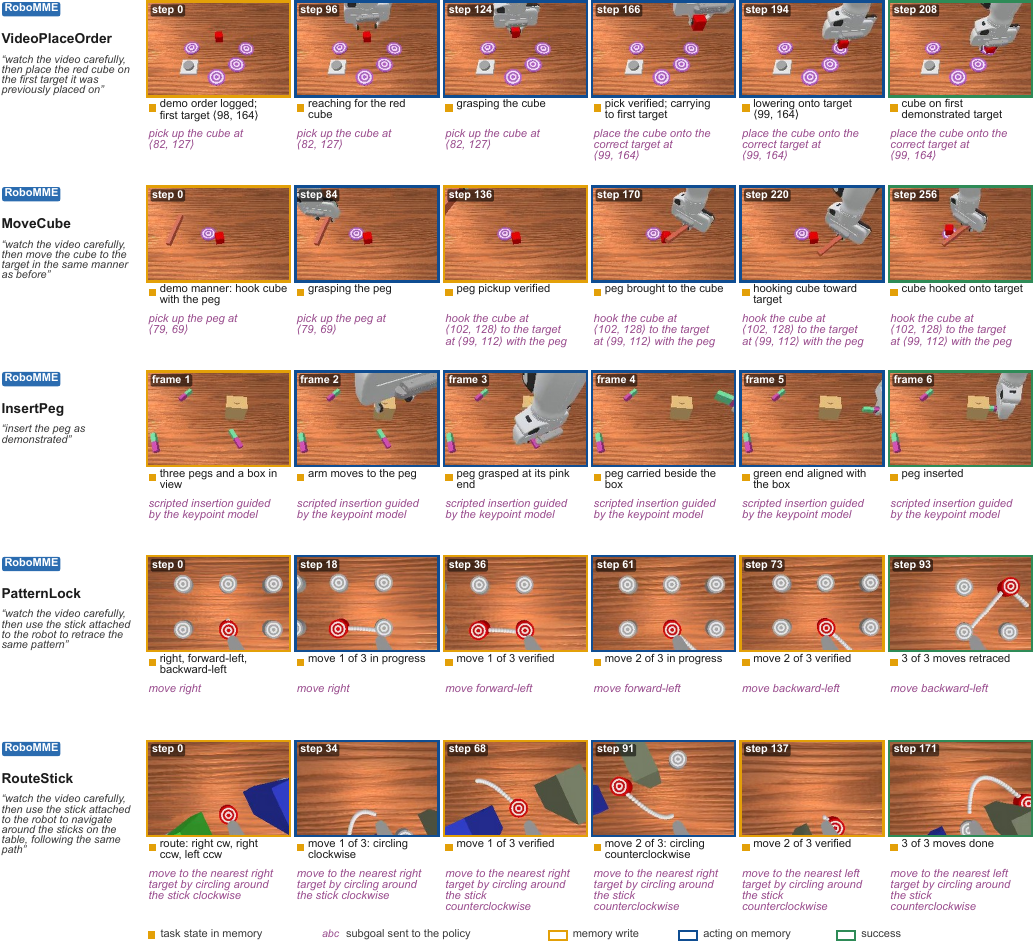}
    \caption{\textbf{\robomme rollouts (3/3)}: VideoPlaceOrder and the Imitation suite (wrist camera for PatternLock and RouteStick).}
    \label{fig:supp_robomme_c}
\end{figure*}

\begin{figure*}[p]
    \centering
    \includegraphics[width=\linewidth]{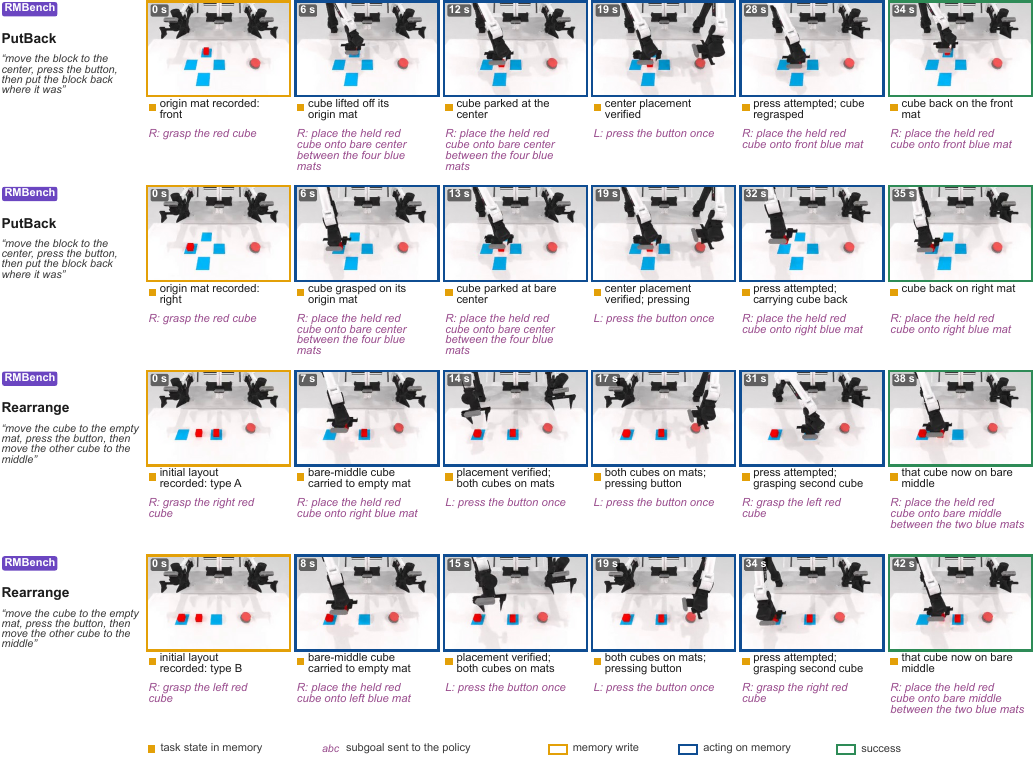}
    \caption{\textbf{\rmbench rollouts (1/2)}: PutBack and Rearrange, two seeds each.}
    \label{fig:supp_rmbench_a}
\end{figure*}

\begin{figure*}[p]
    \centering
    \includegraphics[width=\linewidth]{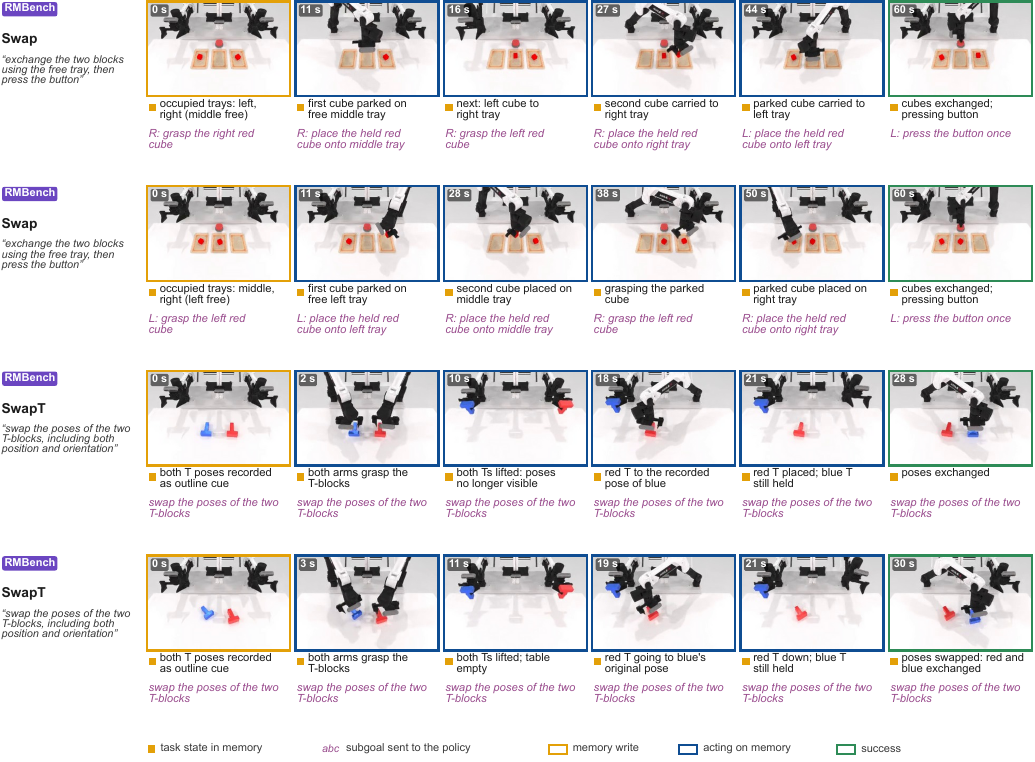}
    \caption{\textbf{\rmbench rollouts (2/2)}: Swap and SwapT, two seeds each. In SwapT the language subgoal is constant and the remembered poses reach the policy as outlines rendered into its image.}
    \label{fig:supp_rmbench_b}
\end{figure*}

\Cref{fig:supp_robomme_a,fig:supp_robomme_b,fig:supp_robomme_c} show one rollout of \method for each of the 16 \robomme tasks, \cref{fig:supp_rmbench_a,fig:supp_rmbench_b} two rollouts for each of four \rmbench tasks, and \cref{fig:supp_real} further real-robot trials. For tasks that begin with a demonstration video, the rollout starts after the demonstration, and the first caption states the facts that the perception programs extracted from it.

%% file: main.bib
@inproceedings{robomme,
  title     = {{RoboMME}: Benchmarking and Understanding Memory for Robotic Generalist Policies},
  author    = {Yinpei Dai and Hongze Fu and Jayjun Lee and Yuejiang Liu and Haoran Zhang and Jianing Yang and Chelsea Finn and Nima Fazeli and Joyce Chai},
  booktitle = {ICML},
  year      = {2026},
  eprint    = {2603.04639}
}

@article{rmbench,
  title   = {{RMBench}: Memory-Dependent Robotic Manipulation Benchmark with Insights into Policy Design},
  author  = {Tianxing Chen and Yuran Wang and Mingleyang Li and Yan Qin and Hao Shi and Zixuan Li and Yifan Hu and Yingsheng Zhang and Kaixuan Wang and Yue Chen and Hongcheng Wang and Junjie Wang and Tianhang Yang and Renjing Xu and Ruihai Wu and Yao Mu and Yaodong Yang and Hao Dong and Ping Luo},
  journal = {arXiv preprint arXiv:2603.01229},
  year    = {2026}
}

@article{simplearm,
  title   = {Simple Agentic Memory for Generalist Robot Policies},
  author  = {Yuyou Zhang and Yunbei Zhang and Miao Li and Janet Wang and Zijian Jin and Shilong Liu and Ding Zhao},
  journal = {arXiv preprint arXiv:2609.36595},
  year    = {2026}
}

@article{ledger,
  title   = {Where Memory Belongs: {Ledger}, an Object Ledger for Memory-Augmented {VLAs}},
  author  = {Tanguy Dieudonn{\'e} and Jack B. Jedlicki and Heng Yang},
  journal = {arXiv preprint arXiv:2609.34554},
  year    = {2026}
}

@inproceedings{memoryvla,
  title     = {{MemoryVLA}: Perceptual-Cognitive Memory in Vision-Language-Action Models for Robotic Manipulation},
  author    = {Hao Shi and Bin Xie and Yingfei Liu and Lin Sun and Fengrong Liu and Tiancai Wang and Erjin Zhou and Haoqiang Fan and Xiangyu Zhang and Gao Huang},
  booktitle = {ICLR},
  year      = {2026},
  eprint    = {2508.19236}
}

@article{evomem,
  title   = {{EvoMem-VLA}: State-Evolution Memory for Long-Horizon Robot Manipulation},
  author  = {Yuheng Na and Zhide Zhong and Junjie He and Junfeng Li and Haodong Yan and Jiaan Wang and Jiaguan Zhu and Yangyang Zheng and Tianyu Huang and Haoang Li},
  journal = {arXiv preprint arXiv:2610.05418},
  year    = {2026}
}

@inproceedings{memer,
  title     = {Scaling up Memory for Robotic Control via Experience Retrieval},
  author    = {Ajay Sridhar and Jennifer Pan and Satvik Sharma and Chelsea Finn},
  booktitle = {ICLR},
  year      = {2026},
  eprint    = {2510.20328}
}

@inproceedings{eventvla,
  title     = {{EventVLA}: Event-Driven Visual Evidence Memory for Long-Horizon Vision-Language-Action Policies},
  author    = {Ganlin Yang and Zhangzheng Tu and Yuqiang Yang and Sitong Mao and Junyi Dong and Tianxing Chen and Jiaqi Peng and Jing Xiong and Jiafei Cao and Jifeng Dai and Wengang Zhou and Yao Mu and Tai Wang},
  booktitle = {CoRL},
  year      = {2026},
  eprint    = {2606.20092}
}

@article{hymes,
  title   = {Skills in Weights, Memory in Code: Hybrid Learning for Memory-Dependent Robot Manipulation},
  author  = {Yunhao Zhao and Zhenyang Ni and Haoyang Chen and Ruohan Zhang and Qi Zhu},
  journal = {arXiv preprint arXiv:2608.09410},
  year    = {2026}
}

@article{robomemarena,
  title   = {{RoboMemArena}: A Comprehensive and Challenging Robotic Memory Benchmark},
  author  = {Huashuo Lei and Wenxuan Song and Huarui Zhang and Jieyuan Pei and Jiayi Chen and Haodong Yan and Han Zhao and Pengxiang Ding and Zhipeng Zhang and Lida Huang and Donglin Wang and Yan Wang and Haoang Li},
  journal = {arXiv preprint arXiv:2605.10921},
  year    = {2026}
}

@inproceedings{notestoself,
  title     = {Notes-to-Self: Scratchpad Augmented {VLAs} for Memory Dependent Manipulation Tasks},
  author    = {Sanjay Haresh and Daniel Dijkman and Apratim Bhattacharyya and Roland Memisevic},
  booktitle = {ICRA},
  year      = {2026},
  eprint    = {2602.21013}
}

@article{simplememvla,
  title   = {{SimpleMemVLA}: A Simple but Effective Native-Video Memory for Vision-Language-Action Models},
  author  = {Cheng Yin and Wang Xu and Junpeng Yang and Sikyuen Tam and Hanyu Liu and Yuan Yao and Xiangrui Zeng and Junbo Cui and Yequan Wang and Zhouping Yin and Yankai Lin},
  journal = {arXiv preprint arXiv:2609.05533},
  year    = {2026}
}

@article{ponderpounce,
  title   = {{PonderPounce}: A Pretrained {MLLM} as an Episode Context Engine for Robot Control},
  author  = {Suhwan Choi and Jaeyoon Jung and Sungkyung Kim and Yunsung Lee and Youngjae Yu},
  journal = {arXiv preprint arXiv:2608.24115},
  year    = {2026}
}

@inproceedings{mikasa,
  title   = {Memory, Benchmark \& Robots: A Benchmark for Solving Complex Tasks with Reinforcement Learning},
  author  = {Egor Cherepanov and Nikita Kachaev and Alexey K. Kovalev and Aleksandr I. Panov},
  booktitle = {ICLR},
  year    = {2026},
  eprint  = {2502.10550}
}

@inproceedings{pi05,
  title   = {{$\pi_{0.5}$}: A Vision-Language-Action Model with Open-World Generalization},
  author  = {Kevin Black and Noah Brown and James Darpinian and Karan Dhabalia and Danny Driess and Adnan Esmail and Michael Equi and Chelsea Finn and Niccolo Fusai and Manuel Y. Galliker and Dibya Ghosh and Lachy Groom and Karol Hausman and Brian Ichter and Szymon Jakubczak and Tim Jones and Liyiming Ke and Devin LeBlanc and Sergey Levine and Adrian Li-Bell and Mohith Mothukuri and Suraj Nair and Karl Pertsch and Allen Z. Ren and Lucy Xiaoyang Shi and Laura Smith and Jost Tobias Springenberg and Kyle Stachowicz and James Tanner and Quan Vuong and Homer Walke and Anna Walling and Haohuan Wang and Lili Yu and Ury Zhilinsky},
  booktitle = {CoRL},
  pages   = {17--40},
  year    = {2025}
}

@inproceedings{pi0,
  title     = {{$\pi_0$}: A Vision-Language-Action Flow Model for General Robot Control},
  author    = {Kevin Black and Noah Brown and Danny Driess and Adnan Esmail and Michael Equi and Chelsea Finn and Niccolo Fusai and Lachy Groom and Karol Hausman and Brian Ichter and Szymon Jakubczak and Tim Jones and Liyiming Ke and Sergey Levine and Adrian Li-Bell and Mohith Mothukuri and Suraj Nair and Karl Pertsch and Lucy Xiaoyang Shi and Laura Smith and James Tanner and Quan Vuong and Anna Walling and Haohuan Wang and Ury Zhilinsky},
  booktitle = {RSS},
  year      = {2025},
  eprint    = {2410.24164}
}

@inproceedings{openvla,
  title   = {{OpenVLA}: An Open-Source Vision-Language-Action Model},
  author  = {Moo Jin Kim and Karl Pertsch and Siddharth Karamcheti and Ted Xiao and Ashwin Balakrishna and Suraj Nair and Rafael Rafailov and Ethan Foster and Grace Lam and Pannag Sanketi and Quan Vuong and Thomas Kollar and Benjamin Burchfiel and Russ Tedrake and Dorsa Sadigh and Sergey Levine and Percy Liang and Chelsea Finn},
  booktitle = {CoRL},
  pages   = {2679--2713},
  year    = {2024}
}

@article{qwen3vl,
  title   = {{Qwen3-VL} Technical Report},
  author  = {Shuai Bai and Yuxuan Cai and Ruizhe Chen and Keqin Chen and Xionghui Chen and Zesen Cheng and Lianghao Deng and Wei Ding and Chang Gao and Chunjiang Ge and Wenbin Ge and Zhifang Guo and Qidong Huang and Jie Huang and Fei Huang and Binyuan Hui and Shutong Jiang and Zhaohai Li and Mingsheng Li and Mei Li and Kaixin Li and Zicheng Lin and Junyang Lin and Xuejing Liu and Jiawei Liu and Chenglong Liu and Yang Liu and Dayiheng Liu and Shixuan Liu and Dunjie Lu and Ruilin Luo and Chenxu Lv and Rui Men and Lingchen Meng and Xuancheng Ren and Xingzhang Ren and Sibo Song and Yuchong Sun and Jun Tang and Jianhong Tu and Jianqiang Wan and Peng Wang and Pengfei Wang and Qiuyue Wang and Yuxuan Wang and Tianbao Xie and Yiheng Xu and Haiyang Xu and Jin Xu and Zhibo Yang and Mingkun Yang and Jianxin Yang and An Yang and Bowen Yu and Fei Zhang and Hang Zhang and Xi Zhang and Bo Zheng and Humen Zhong and Jingren Zhou and Fan Zhou and Jing Zhou and Yuanzhi Zhu and Ke Zhu},
  journal = {arXiv preprint arXiv:2511.21631},
  year    = {2025}
}

@inproceedings{resnet,
  title     = {Deep Residual Learning for Image Recognition},
  author    = {Kaiming He and Xiangyu Zhang and Shaoqing Ren and Jian Sun},
  booktitle = {CVPR},
  pages     = {770--778},
  year      = {2016}
}

@inproceedings{gello,
  title   = {{GELLO}: A General, Low-Cost, and Intuitive Teleoperation Framework for Robot Manipulators},
  author  = {Philipp Wu and Yide Shentu and Zhongke Yi and Xingyu Lin and Pieter Abbeel},
  booktitle = {IROS},
  pages   = {12156--12163},
  year    = {2024}
}

@inproceedings{sayCan,
  title   = {Do As {I} Can, Not As {I} Say: Grounding Language in Robotic Affordances},
  author  = {Brian Ichter and Anthony Brohan and Yevgen Chebotar and Chelsea Finn and Karol Hausman and Alexander Herzog and Daniel Ho and Julian Ibarz and Alex Irpan and Eric Jang and Ryan Julian and Dmitry Kalashnikov and Sergey Levine and Yao Lu and Carolina Parada and Kanishka Rao and Pierre Sermanet and Alexander T. Toshev and Vincent Vanhoucke and Fei Xia and Ted Xiao and Peng Xu and Mengyuan Yan and Noah Brown and Michael Ahn and Omar Cortes and Nicolas Sievers and Clayton Tan and Sichun Xu and Diego Reyes and Jarek Rettinghouse and Jornell Quiambao and Peter Pastor and Linda Luu and Kuang-Huei Lee and Yuheng Kuang and Sally Jesmonth and Nikhil J. Joshi and Kyle Jeffrey and Rosario {Jauregui Ruano} and Jasmine Hsu and Keerthana Gopalakrishnan and Byron David and Andy Zeng and Chuyuan Kelly Fu},
  booktitle = {CoRL},
  pages   = {287--318},
  year    = {2022}
}

@inproceedings{codeaspolicies,
  title   = {Code as Policies: Language Model Programs for Embodied Control},
  author  = {Jacky Liang and Wenlong Huang and Fei Xia and Peng Xu and Karol Hausman and Brian Ichter and Pete Florence and Andy Zeng},
  booktitle = {ICRA},
  pages   = {9493--9500},
  year    = {2023}
}

@inproceedings{innermonologue,
  title   = {Inner Monologue: Embodied Reasoning through Planning with Language Models},
  author  = {Wenlong Huang and Fei Xia and Ted Xiao and Harris Chan and Jacky Liang and Pete Florence and Andy Zeng and Jonathan Tompson and Igor Mordatch and Yevgen Chebotar and Pierre Sermanet and Tomas Jackson and Noah Brown and Linda Luu and Sergey Levine and Karol Hausman and Brian Ichter},
  booktitle = {CoRL},
  pages   = {1769--1782},
  year    = {2022}
}

@inproceedings{trmvla,
  title     = {{TRM-VLA}: Temporal-Aware Chain-of-Thought Reasoning and Memorization for Vision-Language-Action Models},
  author    = {Xiang Li and Ya-Li Li and Yuan Wang and Shengjin Wang},
  booktitle = {CVPR},
  pages     = {10943--10953},
  year      = {2026}
}

@inproceedings{optimusvla,
  title     = {Global Prior Meets Local Consistency: Dual-Memory Augmented Vision-Language-Action Model for Efficient Robotic Manipulation},
  author    = {Zaijing Li and Bing Hu and Rui Shao and Gongwei Chen and Dongmei Jiang and Pengwei Xie and Jianye Hao and Liqiang Nie},
  booktitle = {CVPR},
  pages     = {35135--35145},
  year      = {2026}
}

@inproceedings{hifvla,
  title     = {{HiF-VLA}: Hindsight, Insight and Foresight through Motion Representation for Vision-Language-Action Models},
  author    = {Minghui Lin and Pengxiang Ding and Shu Wang and Zifeng Zhuang and Yang Liu and Xinyang Tong and Wenxuan Song and Shangke Lyu and Siteng Huang and Donglin Wang},
  booktitle = {CVPR},
  pages     = {20732--20742},
  year      = {2026}
}

@inproceedings{fastthinkact,
  title     = {{Fast-ThinkAct}: Efficient Vision-Language-Action Reasoning via Verbalizable Latent Planning},
  author    = {Chi-Pin Huang and Yunze Man and Zhiding Yu and Min-Hung Chen and Jan Kautz and Yu-Chiang Frank Wang and Fu-En Yang},
  booktitle = {CVPR},
  pages     = {5070--5081},
  year      = {2026}
}

@inproceedings{palm,
  title     = {{PALM}: Progress-Aware Policy Learning via Affordance Reasoning for Long-Horizon Robotic Manipulation},
  author    = {Yuanzhe Liu and Jingyuan Zhu and Yuchen Mo and Gen Li and Xu Cao and Jin Jin and Yifan Shen and Zhengyuan Li and Tianjiao Yu and Wenzhen Yuan and Fangqiang Ding and Ismini Lourentzou},
  booktitle = {CVPR},
  pages     = {28096--28110},
  year      = {2026}
}

@inproceedings{foreact,
  title     = {{ForeAct}: Steering Your {VLA} with Efficient Visual Foresight Planning},
  author    = {Zhuoyang Zhang and Shang Yang and Qinghao Hu and Luke J. Huang and James Hou and Yufei Sun and Yao Lu and Song Han},
  booktitle = {CVPR},
  pages     = {37195--37205},
  year      = {2026}
}

@inproceedings{r2vlm,
  title     = {Recurrent Reasoning with Vision-Language Models for Estimating Long-Horizon Embodied Task Progress},
  author    = {Yuelin Zhang and Sijie Cheng and Chen Li and Zongzhao Li and Yuxin Huang and Yang Liu and Wenbing Huang},
  booktitle = {CVPR},
  pages     = {41150--41159},
  year      = {2026}
}

@inproceedings{cotvla,
  title     = {{CoT-VLA}: Visual Chain-of-Thought Reasoning for Vision-Language-Action Models},
  author    = {Qingqing Zhao and Yao Lu and Moo Jin Kim and Zipeng Fu and Zhuoyang Zhang and Yecheng Wu and Zhaoshuo Li and Qianli Ma and Song Han and Chelsea Finn and Ankur Handa and Tsung-Yi Lin and Gordon Wetzstein and Ming-Yu Liu and Donglai Xiang},
  booktitle = {CVPR},
  pages     = {1702--1713},
  year      = {2025}
}

@inproceedings{robobrain,
  title     = {{RoboBrain}: A Unified Brain Model for Robotic Manipulation from Abstract to Concrete},
  author    = {Yuheng Ji and Huajie Tan and Jiayu Shi and Xiaoshuai Hao and Yuan Zhang and Hengyuan Zhang and Pengwei Wang and Mengdi Zhao and Yao Mu and Pengju An and Xinda Xue and Qinghang Su and Huaihai Lyu and Xiaolong Zheng and Jiaming Liu and Zhongyuan Wang and Shanghang Zhang},
  booktitle = {CVPR},
  pages     = {1724--1734},
  year      = {2025}
}

@inproceedings{visprog,
  title     = {Visual Programming: Compositional Visual Reasoning Without Training},
  author    = {Tanmay Gupta and Aniruddha Kembhavi},
  booktitle = {CVPR},
  pages     = {14953--14962},
  year      = {2023}
}

@inproceedings{vipergpt,
  title     = {{ViperGPT}: Visual Inference via {Python} Execution for Reasoning},
  author    = {D{\'\i}dac Sur{\'\i}s and Sachit Menon and Carl Vondrick},
  booktitle = {ICCV},
  pages     = {11888--11898},
  year      = {2023}
}

@inproceedings{lora,
  title     = {{LoRA}: Low-Rank Adaptation of Large Language Models},
  author    = {Edward J. Hu and Yelong Shen and Phillip Wallis and Zeyuan Allen-Zhu and Yuanzhi Li and Shean Wang and Lu Wang and Weizhu Chen},
  booktitle = {ICLR},
  year      = {2022},
  eprint    = {2106.09685}
}

@article{memorywam,
  title   = {{MemoryWAM}: Efficient World Action Modeling with Persistent Memory},
  author  = {Sizhe Yang and Juncheng Mu and Tianming Wei and Chenhao Lu and Xiaofan Li and Linning Xu and Zhengrong Xue and Zhecheng Yuan and Dahua Lin and Jiangmiao Pang and Huazhe Xu},
  journal = {arXiv preprint arXiv:2606.20562},
  year    = {2026}
}

@article{dimwam,
  title   = {{DIM-WAM}: World-Action Modeling with Diverse Historical Event Memory},
  author  = {Kai Wang and Zhaopeng Gu and Yixiang Chen and Yuan Xu and Qisen Ma and Jiabing Yang and Zhaowen Li and Yan Huang and Liang Wang and Peng Su},
  journal = {arXiv preprint arXiv:2606.27677},
  year    = {2026}
}

@inproceedings{rt2,
  title     = {{RT-2}: Vision-Language-Action Models Transfer Web Knowledge to Robotic Control},
  author    = {Brianna Zitkovich and Tianhe Yu and Sichun Xu and Peng Xu and Ted Xiao and Fei Xia and Jialin Wu and Paul Wohlhart and Stefan Welker and Ayzaan Wahid and Quan Vuong and Vincent Vanhoucke and Huong Tran and Radu Soricut and Anikait Singh and Jaspiar Singh and Pierre Sermanet and Pannag R. Sanketi and Grecia Salazar and Michael S. Ryoo and Krista Reymann and Kanishka Rao and Karl Pertsch and Igor Mordatch and Henryk Michalewski and Yao Lu and Sergey Levine and Lisa Lee and Tsang-Wei Edward Lee and Isabel Leal and Yuheng Kuang and Dmitry Kalashnikov and Ryan Julian and Nikhil J. Joshi and Alex Irpan and Brian Ichter and Jasmine Hsu and Alexander Herzog and Karol Hausman and Keerthana Gopalakrishnan and Chuyuan Fu and Pete Florence and Chelsea Finn and Kumar Avinava Dubey and Danny Driess and Tianli Ding and Krzysztof Marcin Choromanski and Xi Chen and Yevgen Chebotar and Justice Carbajal and Noah Brown and Anthony Brohan and Montserrat Gonzalez Arenas and Kehang Han},
  booktitle = {CoRL},
  pages     = {2165--2183},
  year      = {2023},
  eprint    = {2307.15818}
}

@inproceedings{octo,
  title     = {{Octo}: An Open-Source Generalist Robot Policy},
  author    = {Dibya Ghosh and Homer Rich Walke and Karl Pertsch and Kevin Black and Oier Mees and Sudeep Dasari and Joey Hejna and Tobias Kreiman and Charles Xu and Jianlan Luo and You Liang Tan and Lawrence Yunliang Chen and Quan Vuong and Ted Xiao and Pannag R. Sanketi and Dorsa Sadigh and Chelsea Finn and Sergey Levine},
  booktitle = {RSS},
  year      = {2024},
  eprint    = {2405.12213}
}

@inproceedings{statler,
  title     = {Statler: State-Maintaining Language Models for Embodied Reasoning},
  author    = {Takuma Yoneda and Jiading Fang and Peng Li and Huanyu Zhang and Tianchong Jiang and Shengjie Lin and Ben Picker and David Yunis and Hongyuan Mei and Matthew R. Walter},
  booktitle = {ICRA},
  year      = {2024}
}

@inproceedings{successvqa,
  title     = {Vision-Language Models as Success Detectors},
  author    = {Yuqing Du and Ksenia Konyushkova and Misha Denil and Akhil Raju and Jessica Landon and Felix Hill and Nando de Freitas and Serkan Cabi},
  booktitle = {CoLLAs},
  pages     = {120--136},
  year      = {2023}
}

@article{doremi,
  title   = {{DoReMi}: Grounding Language Model by Detecting and Recovering from Plan-Execution Misalignment},
  author  = {Yanjiang Guo and Yen-Jen Wang and Lihan Zha and Zheyuan Jiang and Jianyu Chen},
  journal = {arXiv preprint arXiv:2307.00329},
  year    = {2023}
}

@inproceedings{reflect,
  title     = {{REFLECT}: Summarizing Robot Experiences for Failure Explanation and Correction},
  author    = {Zeyi Liu and Arpit Bahety and Shuran Song},
  booktitle = {CoRL},
  pages     = {3468--3484},
  year      = {2023}
}

@inproceedings{codeasmonitor,
  title     = {Code-as-Monitor: Constraint-aware Visual Programming for Reactive and Proactive Robotic Failure Detection},
  author    = {Enshen Zhou and Qi Su and Cheng Chi and Zhizheng Zhang and Zhongyuan Wang and Tiejun Huang and Lu Sheng and He Wang},
  booktitle = {CVPR},
  pages     = {6919--6929},
  year      = {2025}
}

@article{threet,
  title   = {Experiences with an Architecture for Intelligent, Reactive Agents},
  author  = {R. Peter Bonasso and R. James Firby and Erann Gat and David Kortenkamp and David P. Miller and Marc G. Slack},
  journal = {Journal of Experimental \& Theoretical Artificial Intelligence},
  volume  = {9},
  number  = {2--3},
  pages   = {237--256},
  year    = {1997}
}
